\documentclass{article}

\PassOptionsToPackage{numbers, compress}{natbib}
\usepackage[main, final]{neurips_2026}
\usepackage{graphicx}
\usepackage{tabularx}
\usepackage{booktabs}
\usepackage{array}
\usepackage{longtable}
\usepackage[utf8]{inputenc}
\usepackage[T1]{fontenc}
\usepackage{hyperref}
\usepackage{amsmath}
\usepackage{url}
\usepackage{amsfonts}
\usepackage{nicefrac}
\usepackage{microtype}
\usepackage{xcolor}
\usepackage{tcolorbox}

\title{WuYuEval: A Multi-Level Benchmark for Large Language Models in Solid Waste Management}
\author{\normalfont
\parbox{0.95\textwidth}{\centering
\small
Yi Zhang$^{1,*}$ \quad Hongyang Wang$^{2,*}$ \quad Zheng Hao Leong$^{1}$ \quad Zihao Wu$^{1}$ \\
Kaijun Lin$^{1}$ \quad Zhixing Pan$^{1}$ \quad Qixun Huangfu$^{3}$ \quad Wei Ren$^{4}$ \\
Wenyan Wu$^{5}$ \quad Fangyun Wang$^{6}$ \quad Wenting Yu$^{1}$ \quad Hengyu Lin$^{1}$ \\
Muling Yang$^{7}$ \quad Zongguo Wen$^{1,\dagger}$ \\[0.45em]
\footnotesize
$^{1}$School of Environment, Tsinghua University, Beijing, China. \\
$^{2}$Fudan University, Shanghai, China. \quad
$^{3}$Nanjing University, Nanjing, China. \\
$^{4}$Capital Normal University, Beijing, China. \quad
$^{5}$Tencent Technology (Shenzhen) Company Limited, Shenzhen, China. \\
$^{6}$China University of Petroleum (Beijing), Beijing, China. \quad
$^{7}$Tanwei College, Tsinghua University, Beijing, China. \\
$^{*}$Equal contribution. $^{\dagger}$Corresponding author: \texttt{wenzg@tsinghua.edu.cn} (Zongguo Wen).
}
}

\begin{document}

\maketitle

\begin{abstract}
Large language models (LLMs) are increasingly used as technical assistants, but their competence in solid waste management (SWM) remains difficult to assess because existing benchmarks emphasize general knowledge rather than professional decisions under engineering, environmental, and policy constraints. We introduce WuYuEval, a multi-level benchmark for evaluating LLMs in SWM across foundational knowledge, domain reasoning, and expert decision-making. After quality auditing, WuYuEval contains a Foundation Module with 4,590 closed-ended multiple-choice questions across six task types and eight domain categories, together with an Expert Module with 247 scenario-based open-ended questions involving multi-objective optimization, constraint trade-offs, and system design. For expert tasks, we combine anchor-calibrated LLM-as-a-Judge scoring with Elo-based pairwise comparison. Across 33 LLMs, performance varied widely. The leading model reached 94.64\% accuracy on the Foundation Module, but average accuracy still fell from 84.14\% on easy questions to 42.50\% on hard questions, with lower performance concentrated in calculation, experimental design, urban planning, and open-ended expert tasks. Reasoning-oriented Thinking modes improve most matched model pairs after auditing, but the gains depend on baseline capability and are not uniformly positive. These results suggest that visible deliberation helps only when it remains anchored to units, assumptions, and engineering constraints; otherwise, it may drift from decisive answer boundaries. WuYuEval therefore provides both an evaluation resource and an empirical basis for developing SWM-oriented foundation models with professional reasoning chains and explicit constraint control.
\end{abstract}

\section{Introduction}
\label{sec:intro}

\begin{figure}
    \centering
    \includegraphics[width=1\linewidth]{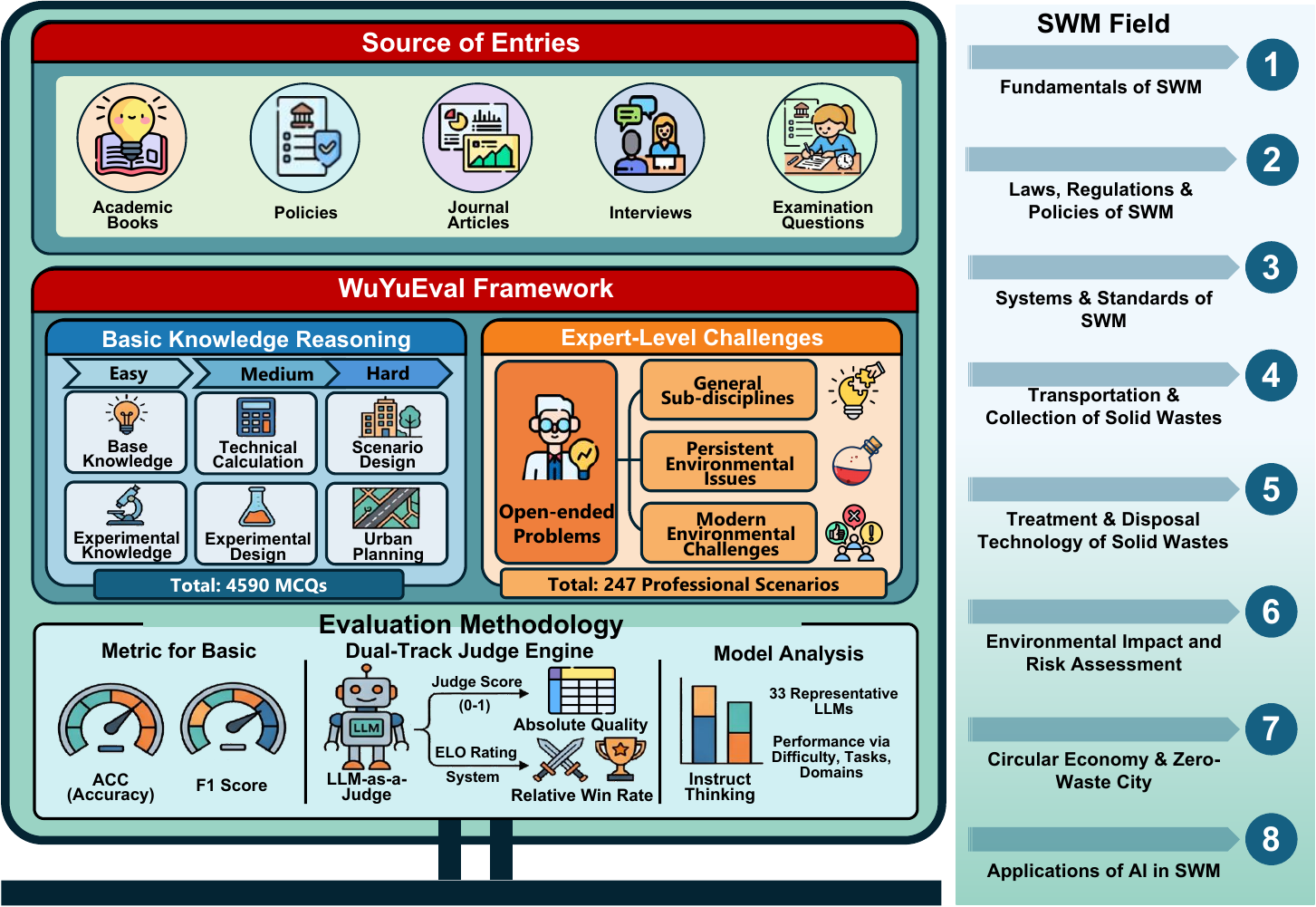}
    \caption{Overview of WuYuEval}
    \label{fig:intro}
\end{figure}

Large language models (LLMs) have rapidly advanced in knowledge storage, retrieval, and generation, enabling their use in specialized domains such as medicine, law, and finance. Beyond factual knowledge, reasoning ability has become a key criterion for assessing the practical utility of LLMs. With techniques such as chain-of-thought prompting \citep{wei2022chain}, LLMs can decompose complex problems into intermediate reasoning steps and have shown strong potential in mathematical reasoning, code generation, and complex decision-making tasks. For scientific and engineering applications, however, models must not only recall domain knowledge but also perform analysis, calculation, design, and decision-making under realistic constraints. This creates a growing need for benchmarks that evaluate both domain knowledge and higher-order reasoning abilities.

Existing general-purpose benchmarks provide important foundations for evaluating LLM capabilities. MMLU \citep{mmlu} measures multitask knowledge across 57 subjects, while BIG-bench \citep{bigbench} evaluates models over more than 200 diverse tasks. Although these benchmarks are useful for assessing general language understanding and reasoning, they are not designed for highly specialized engineering domains. Their tasks are often dominated by closed-form questions, multiple-choice formats, or short-answer evaluation, making them insufficient for measuring multi-objective trade-offs, engineering analysis, system-level planning, and open-ended solution design.

Recent efforts have begun to develop environmental-domain benchmarks. ELLE \citep{elle} provides 1,130 question-answer pairs covering 16 ecological and environmental topics, while EnviroExam \citep{enviroexam} includes 936 questions derived from 42 environmental science courses across undergraduate, master's, and doctoral levels. These benchmarks represent an important step toward domain-specific evaluation in environmental science. However, they still mainly focus on factual question answering, curriculum-based knowledge assessment, and topic-level understanding. Their ability to evaluate multi-step quantitative reasoning, engineering design, complex constraint satisfaction, and system-level decision-making remains limited.

This limitation is particularly critical in solid waste management (SWM), a highly interdisciplinary and practice-oriented field. SWM problems involve waste characterization, treatment and disposal technology selection, pollution control, resource recovery, life-cycle impacts, economic costs, infrastructure conditions, and regulatory constraints. Unlike general environmental QA, SWM tasks require models to conduct professional calculations, compare engineering alternatives, design feasible processes, and balance technical, environmental, economic, and policy objectives. Therefore, evaluating LLMs in SWM requires a benchmark that goes beyond isolated knowledge questions and captures both foundational expertise and expert-level decision-making.

To address this gap, we propose WuYuEval, a hierarchical benchmark for evaluating LLMs in solid waste management. The name ``WuYu'' is inspired by the classical Chinese phrase \emph{Da Fang Wu Yu}, emphasizing an unbounded and systematic perspective. WuYuEval consists of two modules. The first is a foundational capability module containing 4,590 audited closed-ended multiple-choice questions across eight SWM-related areas, covering basic knowledge, professional calculation, scenario analysis, and experimental design. The second is an expert decision-making module containing 247 real-world complex scenario questions, focusing on multi-objective optimization, policy integration, and system-level solution design under complex constraints.

For the expert module, we introduce a two-dimensional evaluation framework. First, an anchor-calibrated LLM-as-a-Judge mechanism scores model outputs in terms of decision quality, reasoning completeness, and solution rationality. Second, an Elo-based adversarial comparison mechanism ranks models through pairwise competition on the same tasks. Together, these evaluation strategies provide both absolute and relative assessments of model performance in expert-level SWM scenarios.

Our contributions are summarized as follows:

\begin{itemize}
    \item We introduce WuYuEval, to our knowledge the first systematic benchmark for LLM evaluation in solid waste management, filling a gap in this specialized, policy-dependent, and practice-oriented domain.

    \item We design a hierarchical evaluation framework covering both foundational and expert-level capabilities across six core task types and eight professional categories.

    \item We provide empirical evaluation and analysis of mainstream LLMs, revealing limitations in multi-step reasoning, consistency control, interdisciplinary integration, and system-level planning.
\end{itemize}

\section{Related Work}
\label{sec:related_work}

\subsection{General LLM Benchmarks}
\label{subsec:general_llm_benchmarks}

A wide range of general-purpose benchmarks have been proposed to evaluate the knowledge, language understanding, and reasoning abilities of large language models (LLMs). Representative examples include MMLU \citep{mmlu}, which measures multitask knowledge across 57 subjects, and BIG-bench \citep{bigbench}, which collects over 200 diverse tasks to probe language, logic, and reasoning capabilities. Other commonly used benchmarks, such as ARC \citep{arc}, PIQA \citep{piqa}, HellaSwag \citep{hellaswag}, and OpenBookQA \citep{openbookqa}, further evaluate commonsense reasoning, physical reasoning, and science question answering. More comprehensive frameworks, such as HELM \citep{helm} and AGIEval \citep{agieval}, emphasize broader and more standardized evaluation of foundation models.

Despite their importance, these benchmarks mainly focus on open-domain or standardized task formats, often relying on multiple-choice accuracy or short-answer correctness. As a result, they provide limited insight into whether LLMs can model complex problems, reason under coupled constraints, or generate feasible decisions in domain-specific engineering scenarios.

Domain-specific benchmarks have also been developed to evaluate specialized knowledge and reasoning. For example, GSM8K \citep{gsm8k} targets mathematical reasoning, LegalBench \citep{legalbench} evaluates legal reasoning, and PubMedQA \citep{pubmedqa} focuses on biomedical question answering. However, most of these benchmarks still emphasize closed-form answers or predefined reasoning paths. They rarely assess process completeness, solution feasibility, or system-level planning under realistic constraints. This limits their applicability to practical scientific and engineering tasks that require multi-step reasoning, cross-process integration, and multi-objective decision-making.

\subsection{Environmental Domain Benchmarks}
\label{subsec:environmental_domain_benchmarks}

Compared with general LLM evaluation, benchmark construction for environmental science remains relatively limited. ELLE \citep{elle} builds 1,130 question-answer pairs covering 16 ecological and environmental topics, providing an evaluation resource for generative AI in the eco-environmental domain. EnviroExam \citep{enviroexam} constructs 936 questions from environmental science curricula across 42 core courses at undergraduate, master's, and doctoral levels. These studies mark an important step toward domain-specific evaluation in environmental science.

However, existing environmental benchmarks are still mainly centered on factual question answering, curriculum-based testing, and topic-level knowledge assessment. They provide limited coverage of higher-order capabilities such as quantitative reasoning, engineering design, experimental planning, and decision-making under complex constraints. This limitation is particularly evident in solid waste management (SWM), where practical tasks require integrated reasoning over waste characteristics, treatment technologies, pollution control, resource recovery, economic costs, infrastructure conditions, and policy constraints.

Therefore, existing benchmarks cannot fully characterize LLMs' capabilities in SWM scenarios that involve professional computation, scenario analysis, process comparison, and system-level decision generation. WuYuEval is designed to fill this gap by providing a hierarchical benchmark for evaluating LLMs from foundational SWM knowledge to expert-level decision-making under realistic constraints.
\section{Dataset Construction}
\label{sec:dataset_construction}

\begin{figure}
    \centering
    \includegraphics[width=1\linewidth]{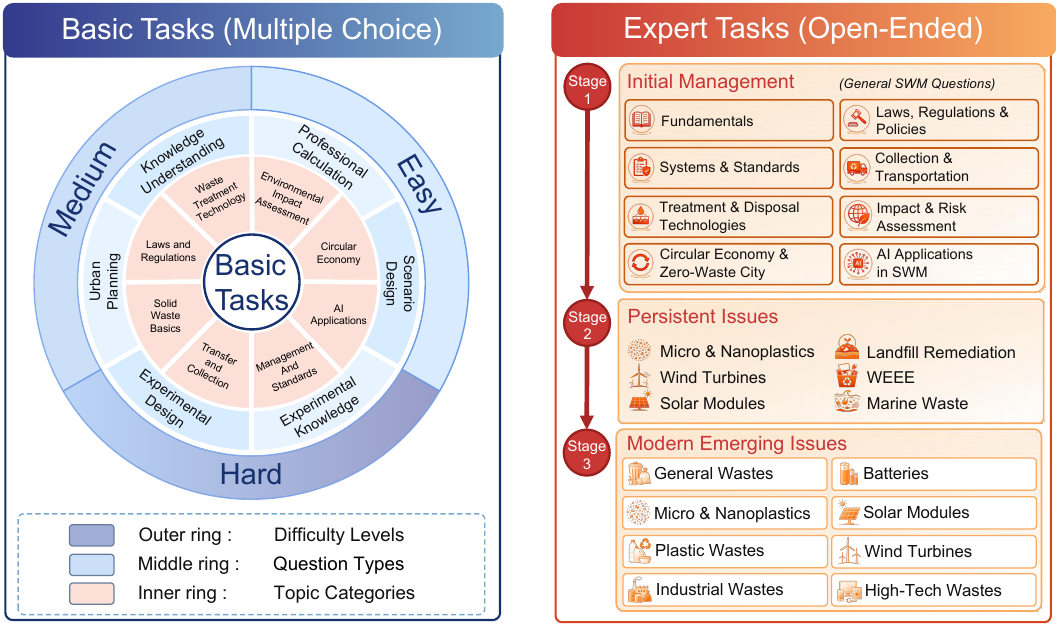}
    \caption{Task design of WuYuEval.}
    \label{fig:task_design}
\end{figure}

\subsection{Task Design and Classification}
\label{subsec:task_design}

WuYuEval is designed as a hierarchical benchmark for evaluating large language models (LLMs) in the solid waste management (SWM) domain. As shown in Figure~\ref{fig:task_design}, WuYuEval consists of two complementary modules: the Foundation Module and the Expert Module. The Foundation Module evaluates models' mastery of standardized SWM knowledge and basic reasoning skills, whereas the Expert Module focuses on open-ended analysis, system-level planning, and decision-making under realistic constraints. This two-level design enables WuYuEval to assess model capabilities from basic knowledge to expert-level problem solving.

The Foundation Module adopts a closed-ended multiple-choice format, including both single-choice and multiple-choice items. Each question has a unique standard answer key, represented either by one option letter or by a set of option letters. Its task taxonomy is organized along three dimensions: task type, topical domain, and difficulty level. First, the module contains six task types: Basic Knowledge, Professional Calculation, Scenario Analysis, Scientific Knowledge, Experimental Design, and Urban Planning. These tasks respectively cover basic concepts and classifications, numerical reasoning and engineering computation, contextual strategy selection, research findings and technical mechanisms, experimental logic, and infrastructure or governance planning.

Second, the Foundation Module covers eight topical domains in SWM: Waste Treatment and Disposal, Environmental Impact and Risk Assessment, Circular Economy and Waste-Free Cities, Management and Standards, Laws and Policies, Solid Waste Fundamentals, AI Applications, and Transfer and Collection. These domains jointly cover both the technical and governance dimensions of SWM, ranging from basic waste classification and treatment technologies to policy compliance, environmental risk, resource circulation, and intelligent management. Third, all foundation questions are divided into three difficulty levels: Easy, Medium, and Hard. Easy questions mainly test direct knowledge recognition, Medium questions require contextual understanding or cross-concept discrimination, and Hard questions involve multi-step reasoning, numerical calculation, or fine-grained distinction among similar technical options.

The Expert Module adopts an open-ended format and is designed for complex SWM scenarios that do not have a single unique answer. These tasks emphasize reasoning completeness, solution feasibility, decision rationality, and system-level coordination. According to the structure of real-world SWM problems, expert questions are divided into three categories: General SWM Sub-domains, Persistent Environmental Issues, and Modern Environmental Challenges. The first category covers SWM fundamentals, laws and policies, circular economy and waste-free cities, and AI applications; the second includes long-term or emerging environmental problems such as micro/nanoplastics, wind turbine waste, photovoltaic module waste, and landfill remediation; and the third focuses on complex waste streams and governance targets, including plastic waste, battery waste, high-tech waste, industrial waste, and municipal solid waste. Together, these tasks require models to integrate technical, environmental, economic, infrastructural, and policy constraints into coherent and implementable solutions.

\subsection{Data Sources and Construction Pipeline}
\label{subsec:data_sources_methods}

\begin{figure}
    \centering
    \includegraphics[width=1\linewidth]{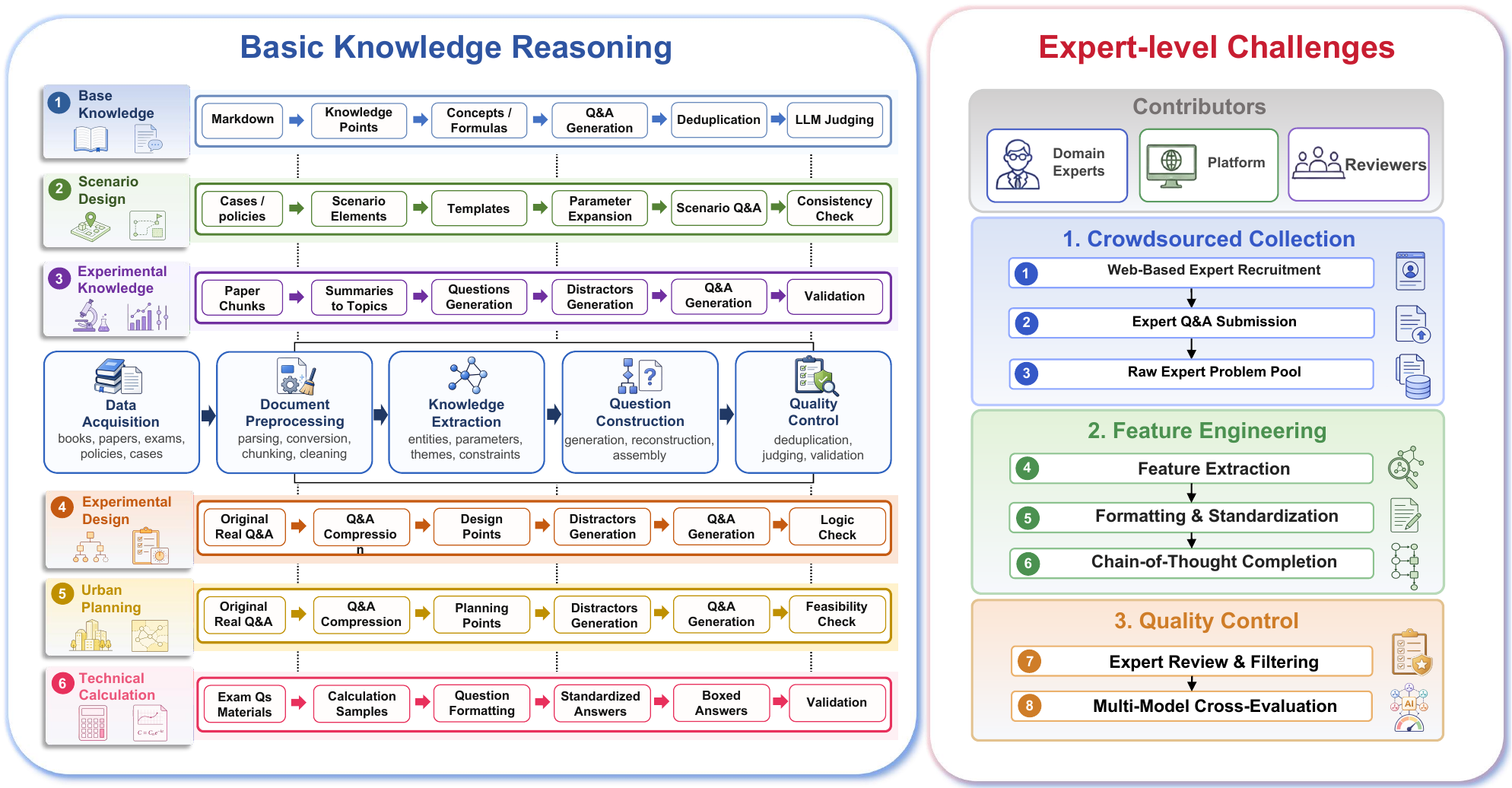}
    \caption{Data construction pipeline of WuYuEval.}
    \label{fig:data_construction}
\end{figure}

The construction of WuYuEval follows a multi-source and multi-stage pipeline, as illustrated in Figure~\ref{fig:data_construction}. The overall goal is to ensure that the benchmark is both scientifically rigorous and practically relevant. Five types of source materials are used: textbooks and teaching materials, policies and regulations, academic papers, expert interviews, and historical exam banks. Textbooks, teaching materials, and exam banks provide standardized concepts, principles, and calculation problems for the Foundation Module. Policies and regulations support tasks related to governance, compliance, standards, and planning. Academic papers contribute recent scientific findings, technical mechanisms, and experimental designs. Expert interviews introduce real-world engineering constraints, practical decision logic, and complex scenarios for expert-level tasks.

For the Foundation Module, the construction process differs according to task type while following a unified quality-control procedure. Basic Knowledge questions are generated from atomic knowledge points extracted from textbooks, standards, and teaching materials. Professional Calculation questions are constructed by extracting calculation kernels from exams and textbook exercises, followed by numerical variable randomization and answer verification. Scenario Analysis questions are produced by combining SWM-related case materials with explicit contextual constraints, such as waste type, treatment objective, infrastructure condition, environmental risk, or policy requirement. Scientific Knowledge questions are derived from academic papers through a generation--evaluation--revision process, which converts research conclusions and technical mechanisms into rigorous multiple-choice questions. Experimental Design questions are built from academic papers, technical reports, and expert knowledge, focusing on objectives, variables, procedures, controls, and causal relationships. Urban Planning questions are constructed from policy reports, planning documents, and expert interviews, emphasizing infrastructure layout, collection and transfer systems, facility planning, and long-term governance strategies.

For the Expert Module, data construction relies more heavily on expert participation. Through the WuYuEval question collection platform, senior SWM experts provide complex engineering problems and reference solutions. The collected materials are then standardized through problem feature extraction, constraint identification, natural language formatting, and reference-solution refinement. Since expert questions are open-ended, the construction process does not aim to produce a single correct answer. Instead, each task is accompanied by reference solution elements and evaluation criteria that support judge-based scoring and pairwise comparison. All expert tasks undergo multiple rounds of review to ensure domain relevance, scenario realism, reasoning depth, and practical feasibility.

Quality control is applied throughout the construction pipeline. For foundation questions, factual correctness, option ambiguity, duplication, answer uniqueness, and metadata consistency are checked through automatic filtering, model-assisted review, and manual auditing. Only questions whose final audit label is \texttt{gold\_ok} with high or medium confidence are retained, yielding 4,590 questions for the reported benchmark and experiments. Calculation questions are verified by symbolic or rule-based computation when sufficient parameters are available. For expert questions, senior experts review problem completeness, constraint rationality, and reference-solution feasibility. We also retain an audit trail for problematic items found during analysis, including under-specified calculation questions; such cases are used to refine the benchmark rather than being attributed solely to model failure. Difficulty labels are calibrated through model-based evaluation and manual verification, and should be interpreted as operational labels relative to the calibration models rather than as intrinsic measures of task hardness.

\subsection{Data Statistics}
\label{subsec:data_statistics}

\begin{figure}[h]
    \centering
    \includegraphics[width=1\linewidth]{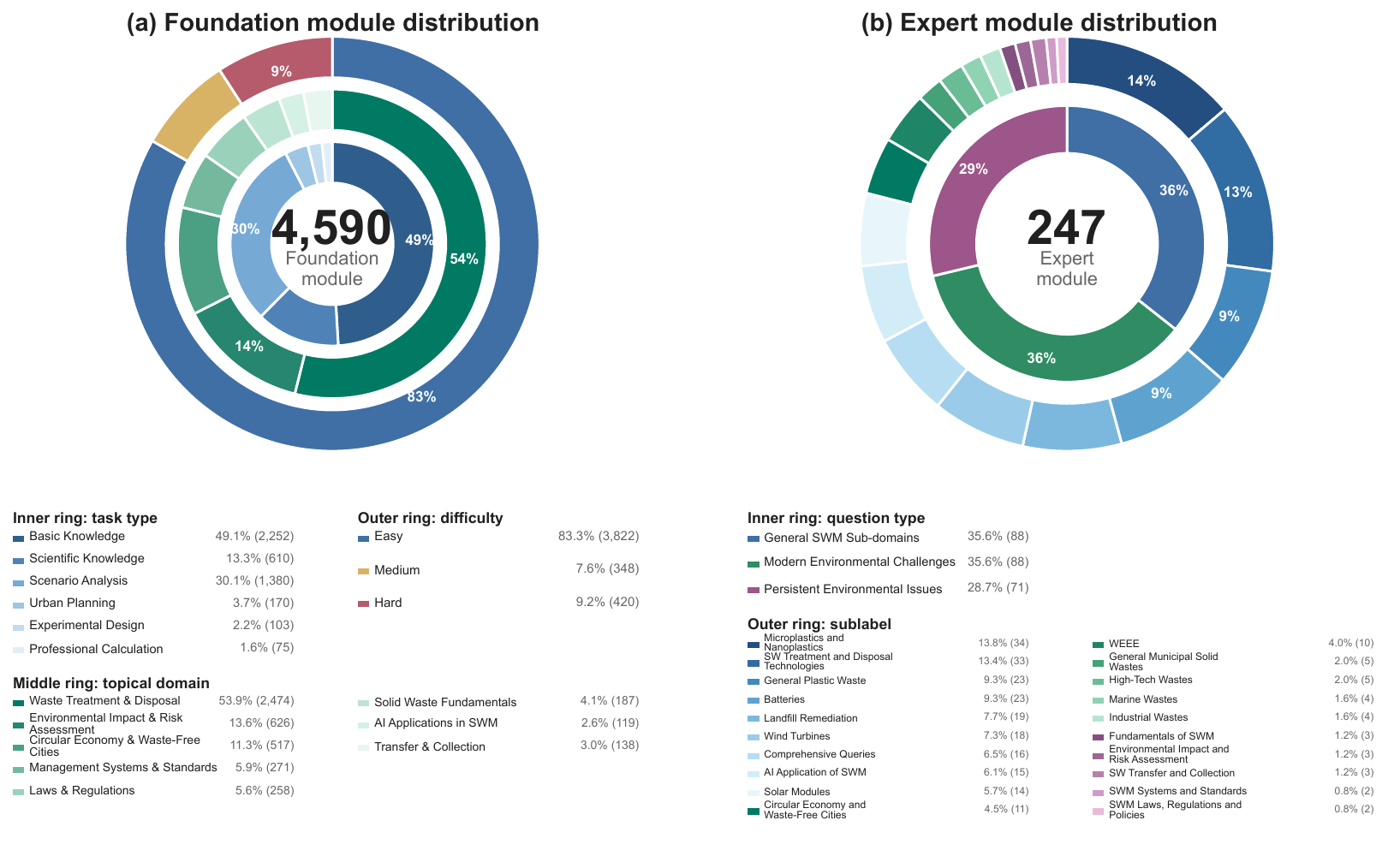}
    \caption{Data statistics of WuYuEval. The Foundation Module is summarized by task type, topical domain, and difficulty level, while the Expert Module is summarized by expert-task type and fine-grained sublabel.}
    \label{fig:data_statistics}
\end{figure}

Figure~\ref{fig:data_statistics} summarizes the overall data distribution of WuYuEval. The benchmark contains 4,837 tasks in total, including 4,590 audited closed-ended multiple-choice questions in the Foundation Module and 247 open-ended questions in the Expert Module. Among the Foundation questions, 4,151 are single-choice items and 439 are multiple-choice items. The Foundation Module consists of questions retained after expert auditing for answer correctness, option ambiguity, metadata consistency, and confidence level. It provides large-scale coverage of standardized SWM knowledge and reasoning tasks, whereas the Expert Module focuses on more complex real-world scenarios that require open-ended analysis, system-level planning, and decision-making.

For the Foundation Module, the distribution is organized along three dimensions: task type, topical domain, and difficulty level. The task-type distribution shows that the module covers both knowledge-oriented questions, such as Basic Knowledge and Scientific Knowledge, and application-oriented questions, such as Scenario Analysis, Experimental Design, Urban Planning, and Professional Calculation. The topical-domain distribution further indicates that WuYuEval covers both technical dimensions, such as treatment and disposal technologies, environmental impact assessment, and transfer and collection, and governance-oriented dimensions, such as circular economy, standards, laws, and policy-related knowledge. The difficulty distribution provides a progressive evaluation structure, ranging from direct knowledge recognition to more challenging questions requiring multi-step reasoning or fine-grained technical discrimination.

For the Expert Module, the distribution is organized by expert-task type and sublabel. The three expert-task types correspond to General SWM Sub-domains, Persistent Environmental Issues, and Modern Environmental Challenges. This structure reflects the dual focus of the Expert Module: it evaluates models not only on general SWM reasoning and governance knowledge, but also on long-standing environmental problems and emerging waste streams. The fine-grained sublabels further cover representative expert-level scenarios, including micro/nanoplastics, photovoltaic modules, wind turbine waste, battery waste, plastic waste, high-tech waste, landfill remediation, and AI-assisted solid waste management.

Overall, the statistical distribution demonstrates the hierarchical design of WuYuEval. The Foundation Module emphasizes broad coverage and standardized assessment, while the Expert Module complements it with smaller-scale but higher-complexity open-ended tasks. Together, the two modules enable evaluation of LLMs from foundational SWM knowledge to expert-level decision-making under realistic constraints.

\section{Results and Analysis}
\label{sec:results}

\subsection{Experimental Evaluation Setup}
\label{subsec:eval_setup}

\subsubsection{Evaluation Metrics}
\label{subsubsec:metrics}

WuYuEval adopts a hybrid evaluation scheme for different task formats. For closed-ended tasks in the Foundation Module, we report Accuracy (ACC) and macro-F1, which are widely used classification metrics \citep{powers2011evaluation}. For open-ended tasks in the Expert Module, we introduce an LLM-as-a-Judge framework inspired by recent studies on scalable LLM-based evaluation and preference judgment \citep{zheng2023judging,liu2023geval}. In addition, we use an Elo-based pairwise comparison mechanism to measure relative model competitiveness \citep{elo1978rating,zheng2023judging}.

\paragraph{Accuracy.}
For single-label closed-ended tasks with unique standard answers, Accuracy is defined as:
\[
ACC=\frac{1}{N}\sum_{i=1}^{N}\mathbb{I}(\hat{y}_{i}=y_{i}),
\]
where $N$ is the number of samples, $\hat{y}_{i}$ is the model prediction, $y_i$ is the ground-truth answer, and $\mathbb{I}(\cdot)$ is the indicator function.

\paragraph{Macro-F1.}
To further measure class-balanced performance for the four answer options, we report the multi-class macro-F1. Let $\mathcal{C}=\{A,B,C,D\}$ denote the option set. For each class $c\in\mathcal{C}$,
\[
P_c=\frac{TP_c}{TP_c+FP_c}, \quad
R_c=\frac{TP_c}{TP_c+FN_c}, \quad
F1_c=\frac{2P_cR_c}{P_c+R_c},
\]
\[
\mathrm{Macro}\text{-}F1=\frac{1}{|\mathcal{C}|}\sum_{c\in\mathcal{C}}F1_c .
\]
When a denominator is zero, the corresponding precision, recall, or F1 term is set to zero. Unlike accuracy, macro-F1 gives equal weight to each answer option and is therefore sensitive to option-level imbalance.

\paragraph{LLM-as-a-Judge weighted score.}
For open-ended expert tasks, where responses cannot be evaluated by a single reference answer, WuYuEval adopts a dual-path LLM-as-a-Judge framework. The first path evaluates the overall quality of a response, while the second path evaluates its structural completeness through an answer graph.

For the overall graded score, Judge Model A compares the model response $a_i$ with the reference answer $g_i$ using a unified rubric and outputs a raw score $r_i\in[0,100]$. To reduce scale drift, each question includes two anchors: a gold answer $r_i^{gold}$ and a null answer $r_i^{null}$. The raw score is normalized as:
\[
s_i^{overall}=\mathrm{clip}\left(\frac{r_i-r_i^{null}}{r_i^{gold}-r_i^{null}},0,1\right).
\]
This score reflects content coverage, relevance, organization, reasoning depth, and rigor. The anchor normalization also explains why the empirical lower bound can be 0.0000 in the result tables.

For structural evaluation, an extraction model first converts response $a_i$ into an answer graph:
\[
G_i=E(a_i)=(V_i,E_i),
\]
where $V_i$ denotes key element nodes and $E_i$ denotes relations among them. Judge Model B then assigns a graph-based structural raw score, which is normalized by the same gold/null anchor procedure:
\[
s_i^{graph}=J(q_i,G_i,g_i), \quad s_i^{graph}\in[0,1].
\]
This score emphasizes key-element coverage, logical organization, framework rationality, and structural consistency with the reference answer. In our implementation, graph extraction uses \texttt{gemini-2.5-flash}, and the overall and graph judge calls use \texttt{deepseek-v3.2-exp} with temperature 0.3. The repair-and-summary script uses compact JSON-only prompts and re-judges abnormal question files when anchors or score fields are missing, or when more than 80\% of model scores are zero. Model identifiers are preserved in the current judge prompts, which remains a limitation for future blinded evaluation.

The final judge score is computed as:
\[
S_i^{judge}=0.7s_i^{overall}+0.3s_i^{graph}.
\]
This weighted fusion balances global response quality and structural reasoning completeness.

\paragraph{Elo rating mechanism.}
For expert-level open-ended tasks, WuYuEval also introduces an Elo-based pairwise comparison mechanism. Given two models $i$ and $j$ with ratings $R_i$ and $R_j$, their expected scores are:
\[
E_i=\frac{1}{1+10^{(R_j-R_i)/400}}, \quad
E_j=\frac{1}{1+10^{(R_i-R_j)/400}}=1-E_i .
\]
After each comparison, ratings are updated as:
\[
R_i'=R_i+K(S_i-E_i),
\]
\[
R_j'=R_j+K(S_j-E_j),
\]
where $S_i\in\{1,0.5,0\}$ denotes win, draw, or loss. All models are initialized at $R=1500$, with $K=32$, $24$, and $16$ for high-, medium-, and low-confidence judgments, respectively. The repaired summary replays the stored \texttt{elo\_matches} in ascending question-index order and contains 2,964 pairwise comparisons, corresponding to 12 matches per expert question. Since pair sampling leaves some models unmatched for a given question, match counts differ across models; Elo is therefore interpreted as a relative ranking under the stored comparison protocol.

\subsubsection{LLM-based Difficulty Classification Strategy}
\label{subsubsec:difficulty_classification}

\begin{figure}[h]
    \centering
    \includegraphics[width=0.9\linewidth]{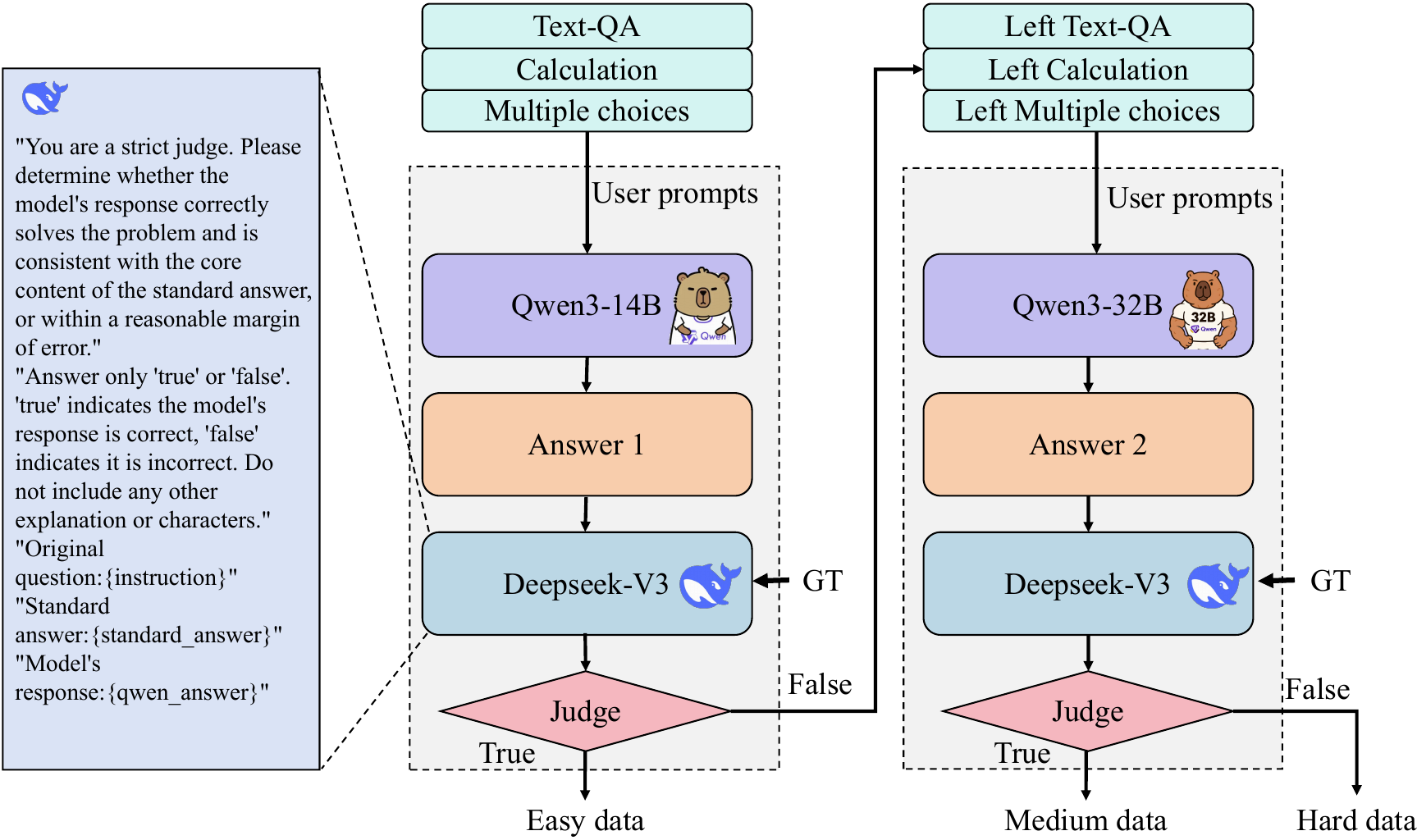}
    \caption{LLM-based difficulty classification strategy.}
    \label{fig:difficulty}
\end{figure}

To automate difficulty grading, we design a multi-model judgment pipeline. Each generated question is first answered by \texttt{Qwen3-14B} \citep{yang2025qwen3}, and the response is judged against the ground truth by \texttt{DeepSeek-V3} \citep{deepseek2024v3}. If the answer is correct, the sample is labeled as Easy. Otherwise, the question is passed to \texttt{Qwen3-32B}. If \texttt{Qwen3-32B} answers correctly, the sample is labeled as Medium; otherwise, it is labeled as Hard. To improve reliability, we further apply dual-layer verification and manual review for the Hard subset. Since Qwen3-family models are also included in the downstream evaluation, these labels should be understood as calibration-model-relative difficulty labels. They are used for diagnostic stratification rather than as model-independent ground truth about intrinsic question difficulty.

\subsubsection{Open-ended QA Evaluation Framework}
\label{subsubsec:open_qa_eval}

\begin{figure}
    \centering
    \includegraphics[width=1\linewidth]{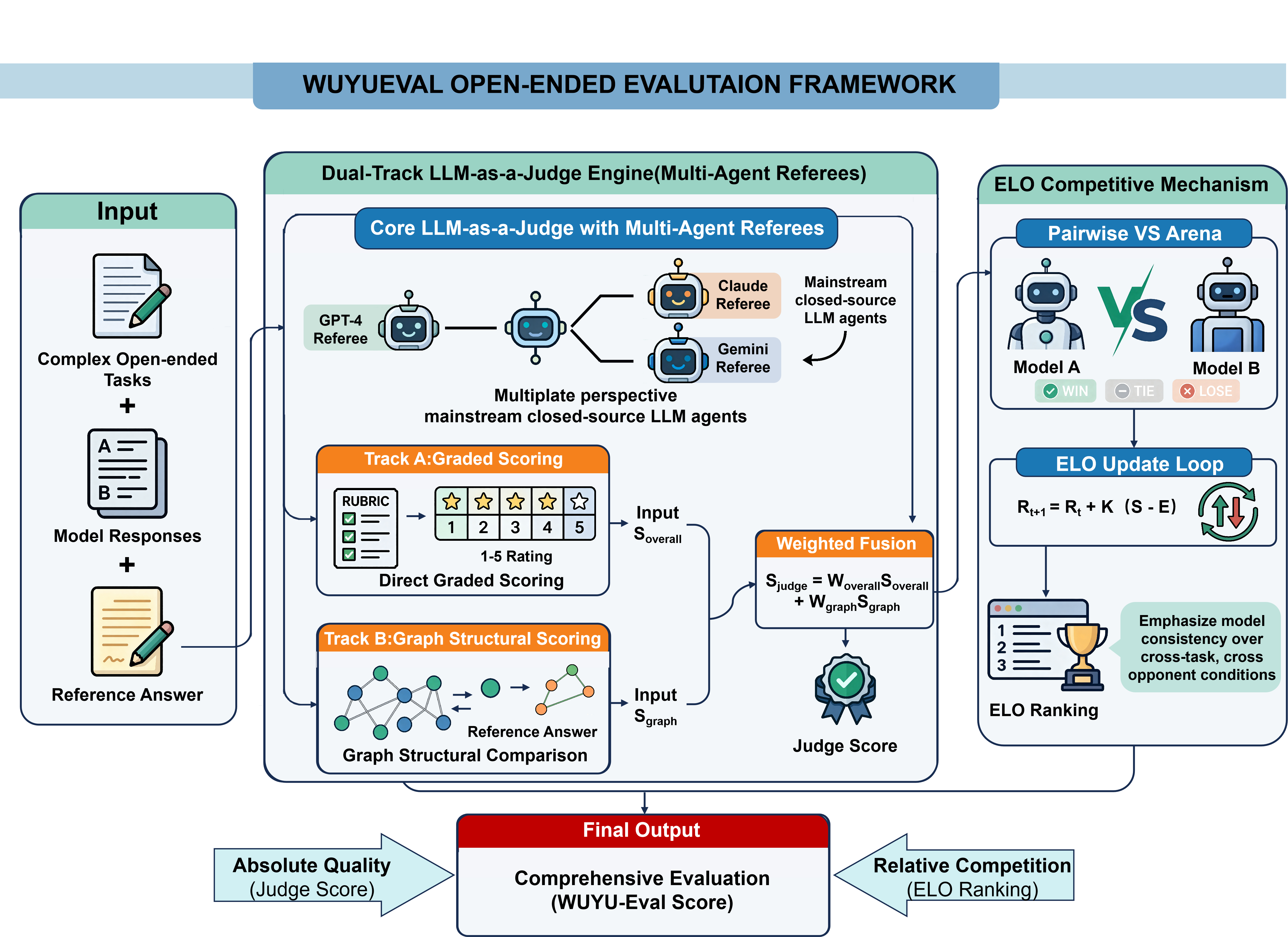}
    \caption{Open-ended QA evaluation framework.}
    \label{fig:metrics}
\end{figure}

To evaluate open-ended tasks such as experimental design, waste-free city planning, and expert-level decision-making, WuYuEval adopts an anchor-calibrated LLM-as-a-Judge mechanism. This design follows recent work showing that strong LLMs can serve as scalable evaluators for open-ended generation tasks \citep{zheng2023judging,liu2023geval}. The evaluation follows two paths: an overall graded score that measures content coverage, relevance, and reasoning quality, and an answer-graph structural score that measures key-element organization and logical completeness. The two paths are fused with weights of 0.7 and 0.3, respectively. The exact prompt templates are reported in Appendix~\ref{app:judge_prompts}.

\subsection{Experimental Results Analysis}
\label{subsec:exp_results}

\subsubsection{Foundational Logical Judgment}
\label{subsubsec:foundational_logic}

We evaluate 33 LLMs on the audited WuYuEval Foundation Module, which contains 4,590 closed-ended multiple-choice questions. The candidate pool also included one task-specific fine-tuned variant, \texttt{Qwen3-14B-sft}, which is excluded from all statistics reported in this paper because it is not a general evaluated baseline. Model performance is analyzed across overall capability, difficulty level, reasoning mode, task type, and SWM domain using ACC and macro-F1.

\begin{figure}[h]
    \centering
    \includegraphics[width=1\linewidth]{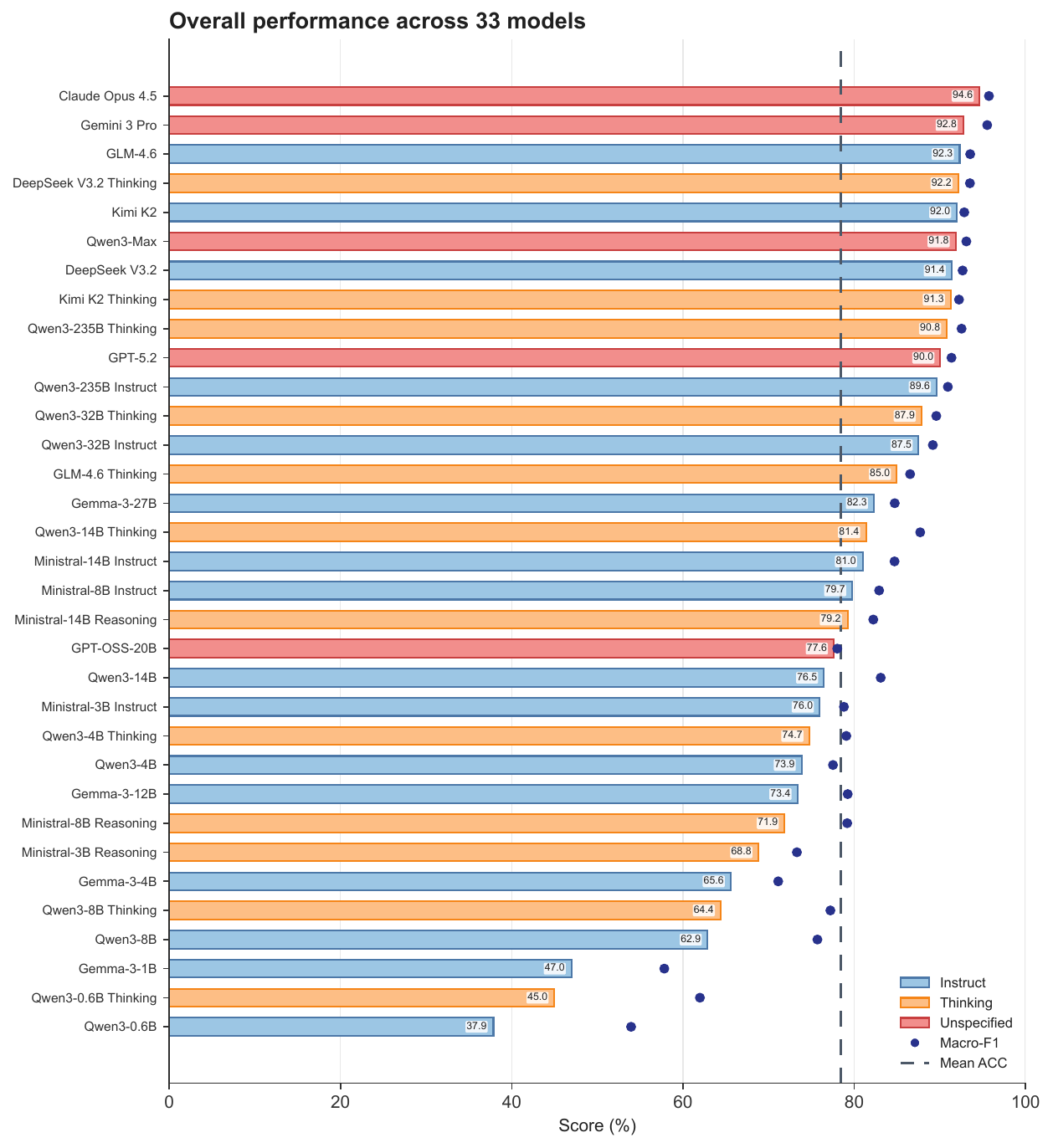}
    \caption{Overall Performance across 33 Models on WuYuEval.}
    \label{fig:overall_performance}
\end{figure}

\paragraph{Overall performance on the basic-question module.}
Figure~\ref{fig:overall_performance} reports the overall performance of 33 models on the audited basic-question module of WuYuEval. Model performance varies substantially: $\mathrm{ACC}$ ranges from 37.86\% to 94.64\%, and macro-F1 ranges from 53.94\% to 95.73\%. The highest $\mathrm{ACC}$ is achieved by \texttt{claude-opus-4.5} at 94.64\%, followed by \texttt{gemini-3-pro-preview}, \texttt{glm-4.6}, \texttt{DeepSeek-V3.2-Thinking}, \texttt{kimi-k2-preview}, and \texttt{qwen3-max}. By contrast, \texttt{Qwen3-0.6B} and \texttt{Qwen3-0.6B-thinking} obtain 37.86\% and 44.99\% ACC, respectively. These results indicate that WuYuEval retains strong discrimination after quality auditing while reducing confounds from ambiguous or under-specified items.

At the distributional level, the average $\mathrm{ACC}$ across the 33 evaluated models is 78.43\%, and the average macro-F1 is 82.96\%. Since macro-F1 is computed over answer-option classes rather than over samples, it reflects option-level balance rather than stability across SWM task categories; task- and domain-level variation is analyzed separately below.

Although the best-performing model achieves an $\mathrm{ACC}$ of 94.64\%, this still corresponds to 246 incorrect answers among 4,590 audited questions. The remaining errors are concentrated in tasks closer to professional workflows, such as scientific inference, engineering calculation, experimental design, and urban governance. Thus, current LLMs appear useful as assistive analytical tools, but not yet reliable as autonomous decision-making systems in SWM.

\paragraph{Performance by question difficulty.}
A further breakdown by question difficulty shows that model performance decreases substantially as task complexity increases. As reported in Table~\ref{tab:difficulty}, the average $\mathrm{ACC}$ on Easy, Medium, and Hard questions is 84.14\%, 59.10\%, and 42.50\%, respectively, while the corresponding macro-F1 scores are 88.30\%, 64.06\%, and 49.98\%. The accuracy gap between Easy and Hard questions reaches 41.64 percentage points, indicating a pronounced degradation of current models on highly complex SWM problems.

Models perform relatively well on Easy questions, which mainly involve direct concept recognition, standardized knowledge, and short reasoning chains. Performance drops sharply on Medium and Hard questions, where models must integrate evidence, distinguish similar technical options, or reason over engineering constraints. Hard questions often reveal a reliance on general domain intuition rather than question-specific evidence; for instance, models may prefer a process that appears environmentally mild even when the provided LCA values support a different answer. This pattern suggests that the bottleneck is not only knowledge coverage, but also reliable use of evidence, assumptions, and system boundaries.

The difficulty analysis also motivates continued dataset auditing. We identified cases in which the visible statement of a calculation question lacks sufficient parameters for a well-posed solution. Such items should be revised or excluded in future releases, and low scores on them should not be attributed solely to model incapability.

\paragraph{Thinking model gain analysis.}
A comparison between standard instruction-following modes and their Thinking-mode counterparts shows that explicit reasoning is beneficial for most matched model pairs after auditing, but the gains remain uneven. Among the nine matched model pairs, seven improve in $\mathrm{ACC}$ after enabling Thinking mode, while only \texttt{GLM-4.6} and \texttt{Kimi-K2} decline. The largest gains appear for \texttt{Qwen3-0.6B} (+7.12 percentage points) and \texttt{Qwen3-14B} (+4.90), whereas the two negative cases occur in already strong large-model families whose standard modes exceed 92\% ACC. This suggests a size- and capability-dependent pattern: weaker or smaller models may benefit from explicit reasoning because it supplies additional intermediate structure, while highly capable models with strong parametric knowledge may sometimes over-deliberate, introduce plausible but unnecessary considerations, or drift from the decisive answer boundary. A task-level breakdown is mixed: Thinking improves the mean ACC on Basic Knowledge, Scientific Knowledge, and Calculation, but decreases it on Scenario Analysis, Urban Planning, and Experimental Design. This pattern does not imply that longer reasoning is automatically reliable. Rather, it suggests that reasoning helps when it remains anchored to domain-specific evidence and constraints. In SWM tasks, the decisive step may be selecting the governing constraint, checking a unit, respecting a regulatory boundary, or identifying the key risk posture.

\begin{figure}[h]
    \centering
    \includegraphics[width=1\linewidth]{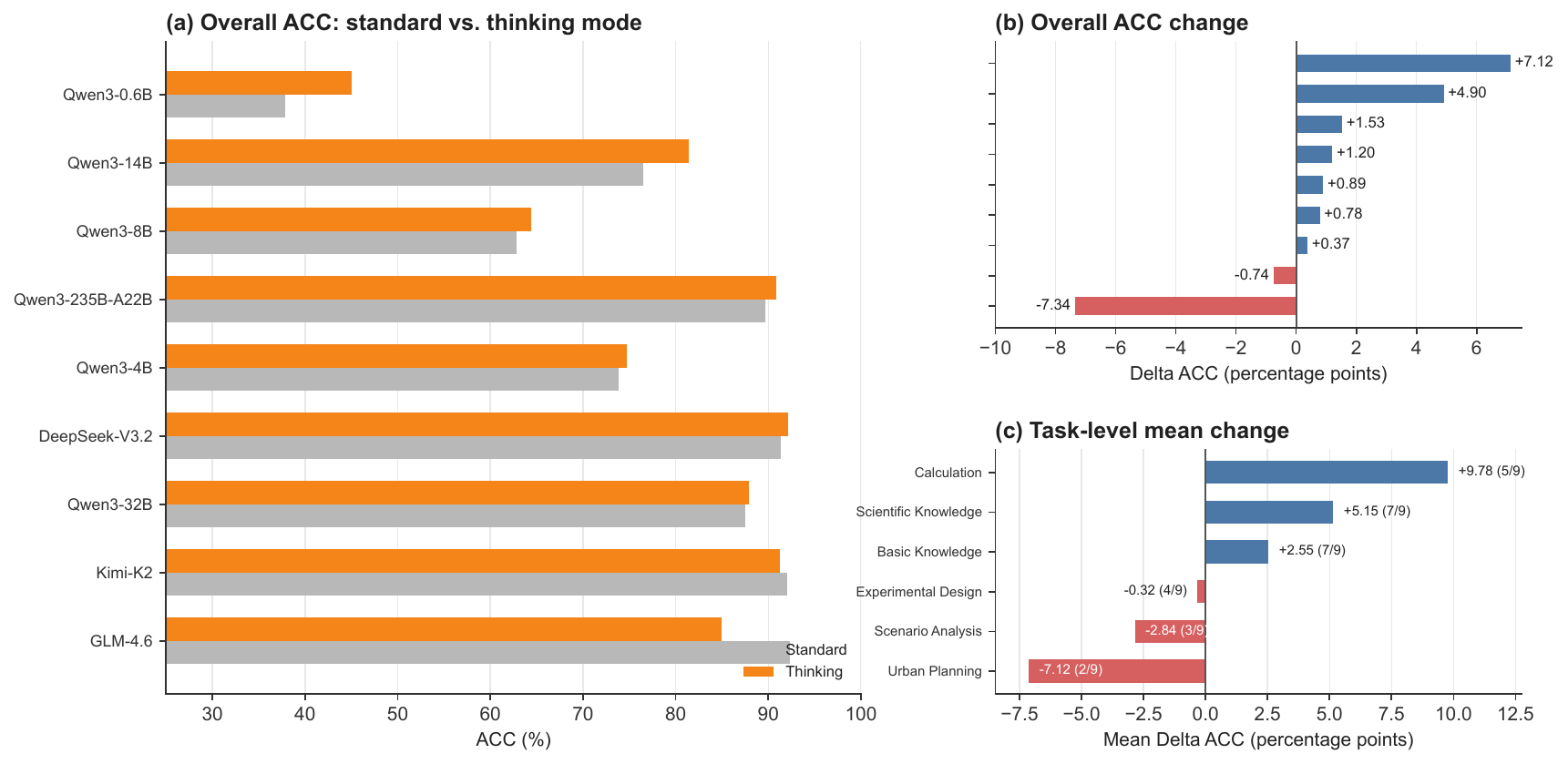}
    \caption{Thinking model gain analysis. Panels (a)--(b) compare overall accuracy for nine standard/Thinking model pairs, while panel (c) summarizes the mean task-level ACC change across the same pairs. Values in parentheses in panel (c) indicate how many of the nine pairs improve on that task.}
    \label{fig:thinking}
\end{figure}

\paragraph{Task- and domain-specific model performance.}

\begin{figure}
    \centering
    \includegraphics[width=1\linewidth]{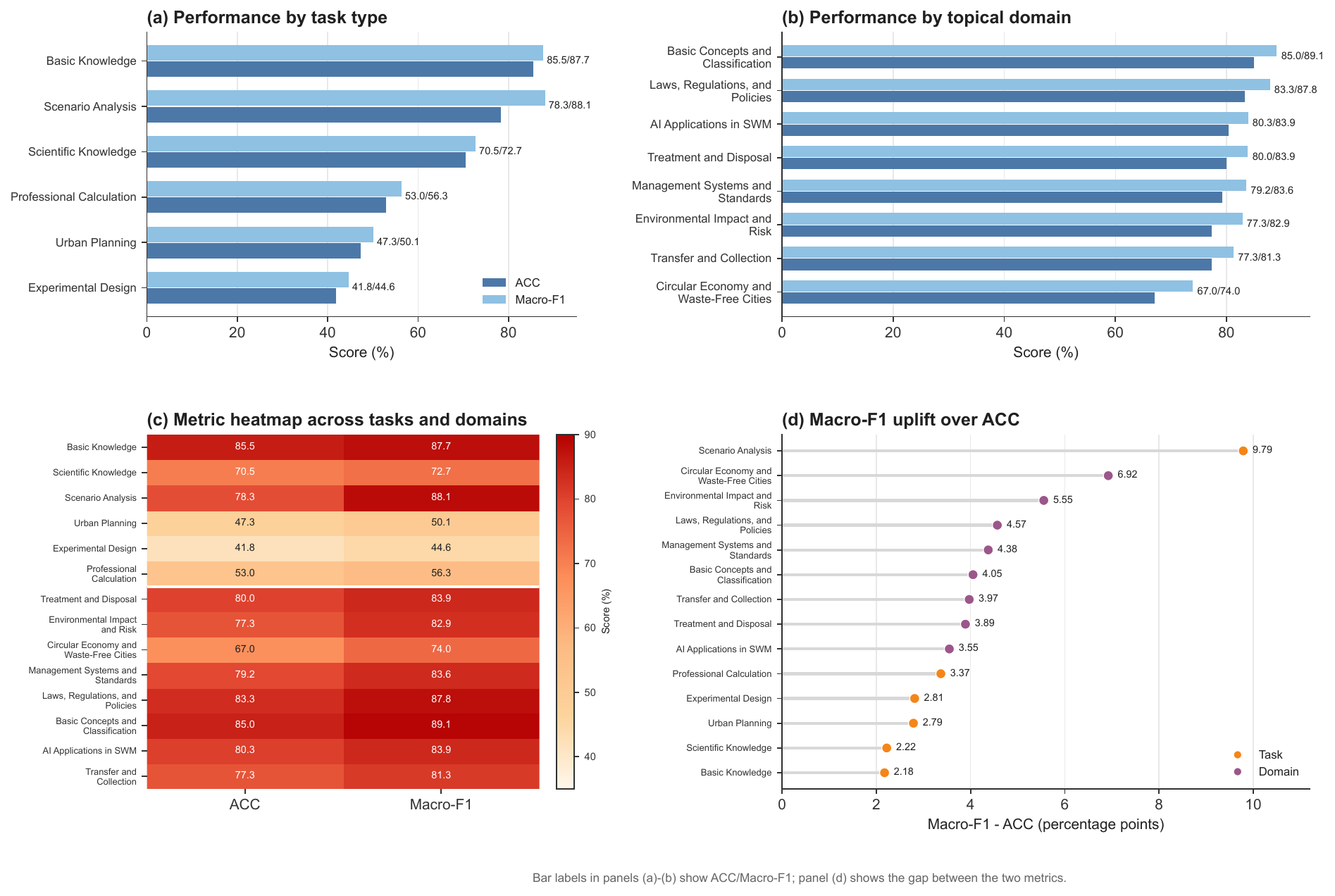}
    \caption{Model performance across task types and SWM domains. Panels (a)--(b) show task-level and domain-level ACC/Macro-F1, while panels (c)--(d) summarize the same patterns through a compact heatmap and a Macro-F1--ACC metric-gap view. Tasks with well-defined answer boundaries (Basic Knowledge, Scenario Analysis) achieve higher scores, while tasks requiring structured reasoning (Experimental Design, Urban Planning, Calculation) show weaker performance. Domains with strong normative or regulatory structure also yield higher performance, whereas systemically complex domains (Circular Economy and Waste-Free Cities, Environmental Impact) reveal greater limitations.}
    \label{fig:ana4}
\end{figure}

We further examine model performance along the dimensions of task type and domain, revealing a pronounced imbalance in capability distribution. As shown in Figures~\ref{fig:ana4}(a)--(b), across the six major task categories, models achieve their highest performance on \textit{Basic Knowledge} and \textit{Scenario Analysis} tasks, with average accuracy (ACC) of 85.52\% and 78.31\%, and macro-F1 scores of 87.69\% and 88.10\%, respectively. In contrast, \textit{Experimental Design}, \textit{Urban Planning}, and \textit{Calculation} tasks show substantially lower performance, with average ACCs of 41.84\%, 47.31\%, and 52.97\%, respectively. Because Calculation and Experimental Design contain only 75 and 103 audited questions, respectively, these results should be interpreted with the sample-size limitations in mind. Even with this caveat, the consistent drop across lower-sample but higher-complexity categories suggests that current models are more adept at tasks emphasizing knowledge retrieval, pattern matching, and conventional contextual understanding, whereas tasks requiring explicit structured reasoning, multi-factor coordination, and solution completeness control expose significant weaknesses.

This discrepancy has clear implications for the solid waste management (SWM) domain. Basic Knowledge and Scenario Analysis tasks primarily involve conceptual, classificatory, regulatory, and standardized knowledge with relatively well-defined answer boundaries. In contrast, research-oriented, computational, planning, and experimental-design tasks more closely resemble real-world SWM problems, where models must jointly consider technical pathways, environmental constraints, resource allocation, and governance objectives.

Figures~\ref{fig:ana4}(c)--(d) summarize the same task and domain patterns through a compact heatmap and metric-gap view. At the domain level, \textit{Basic Concepts and Classification of Solid Waste} achieves high average ACC (85.01\%), while \textit{Solid Waste Laws, Regulations, and Policies} also performs well (83.25\%). Conversely, \textit{Circular Economy and Waste-Free Cities} remains the lowest-scoring large domain (67.04\%), with \textit{Environmental Impact and Risk Assessment of Solid Waste} and \textit{Solid Waste Transfer and Collection} also below the strongest conceptual and regulatory categories. These results suggest that models produce more stable outputs in domains with clear conceptual boundaries, strong normative structure, and well-defined answers, whereas topics involving higher system interconnectivity and coupled objectives reveal greater limitations.

Heatmaps further show that performance differences are not localized to a single task or domain. Normative and deterministic knowledge areas align better with current models' retrieval-oriented strengths, whereas circular economy, waste-free cities, and environmental risk assessment require cross-sector governance, multi-objective trade-offs, and systemic judgment. These results indicate that current models remain less reliable in tasks closer to practical governance and solution design.

\subsubsection{Expert-Level Open-Ended Challenges}
\label{subsubsec:expert_challenges}

\paragraph{LLM-as-a-Judge evaluation.}

\begin{figure}[ht]
    \centering
    \includegraphics[width=\textwidth]{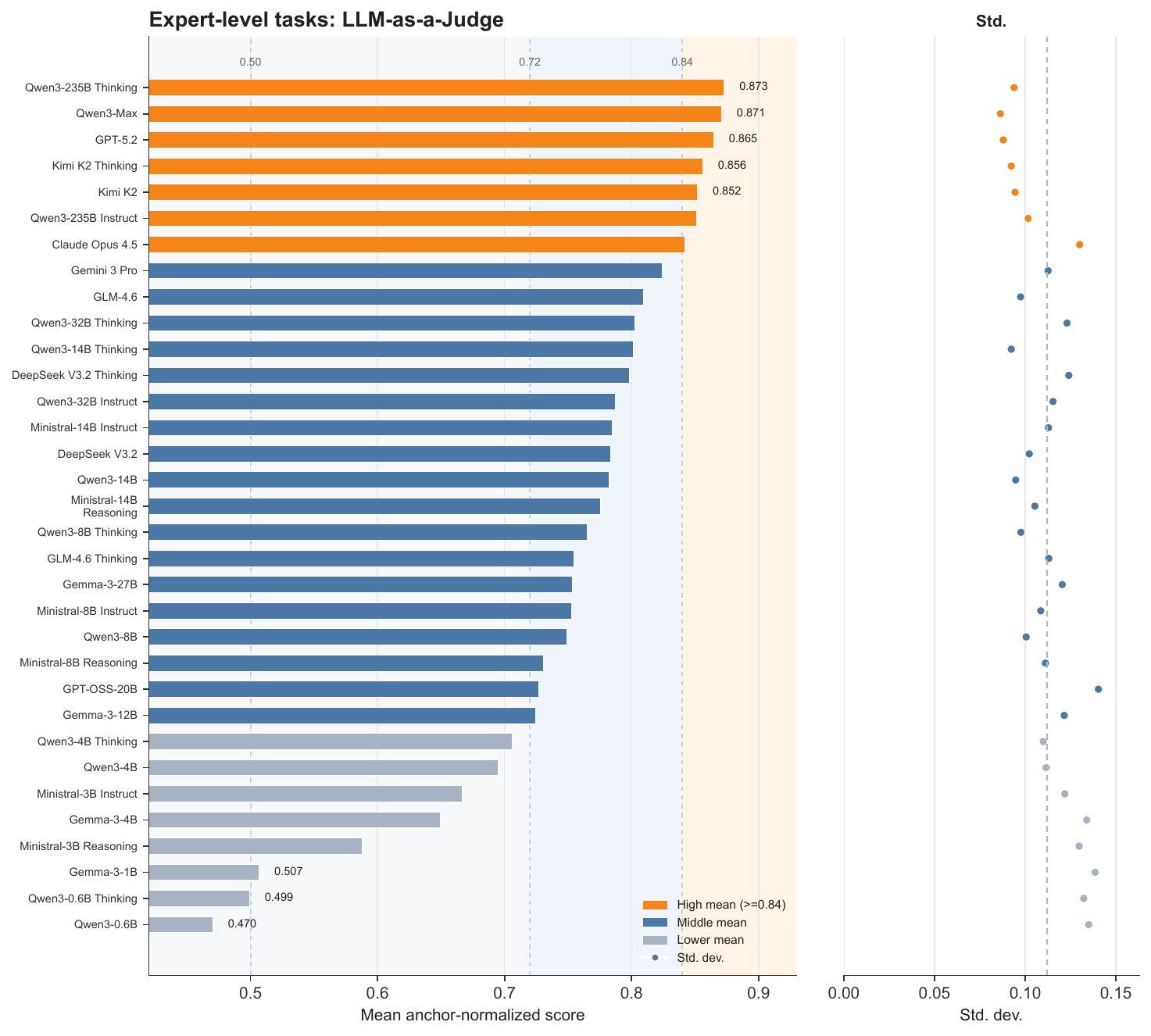}
    \caption{Model performance on expert-level open-ended tasks. The main panel shows mean anchor-normalized judge scores, and the side panel reports the standard deviation of judge scores across expert questions. Top-performing models achieve higher average scores and more structured answers, while lower-performing models display both lower averages and reduced solution completeness.}
    \label{fig:expert_perf}
\end{figure}

We further evaluate models on expert-level open-ended tasks using the LLM-as-a-Judge framework. Figure~\ref{fig:expert_perf} reports anchor-normalized final scores, where the main panel shows mean scores and the side panel reports score standard deviations. Top-performing models achieve average scores above 0.84, while lower-performing models drop to around 0.50, suggesting that capability gaps widen in complex open-ended scenarios. Minimum values of 0.0000 correspond to responses calibrated near the null anchor for specific questions.

Specifically, \texttt{qwen3-235b-a22b-thinking} achieves the highest average score (0.873), followed by \texttt{qwen3-max}, \texttt{gpt-5.2}, \texttt{kimi-k2-thinking}, and \texttt{kimi-k2-preview}, all exceeding 0.85. In contrast, \texttt{Qwen3-0.6B}, \texttt{Qwen3-0.6B-thinking}, and \texttt{gemma-3-1b-it} perform poorly, with mean scores below 0.51. These results indicate that, on expert-level tasks, leading models can generate high-quality, well-structured answers, whereas smaller models still struggle with comprehensive analysis and solution organization.

This pattern suggests that expert SWM tasks demand more than localized knowledge retrieval. They require coordinated reasoning, structural organization, and explicit handling of technical, environmental, economic, and policy constraints. Although leading models can generate relatively complete plans, their non-negligible score variation indicates that robustness across scenarios is still limited.

\paragraph{Elo-based adversarial evaluation.}

To further assess model competitiveness on expert-level open-ended tasks, we introduce an Elo-based adversarial evaluation framework based on pairwise comparisons for the same question. Unlike LLM-as-a-Judge, which estimates absolute answer quality, Elo ranking emphasizes relative performance across opponents and questions.

\begin{figure}[ht]
    \centering
    \includegraphics[width=\textwidth]{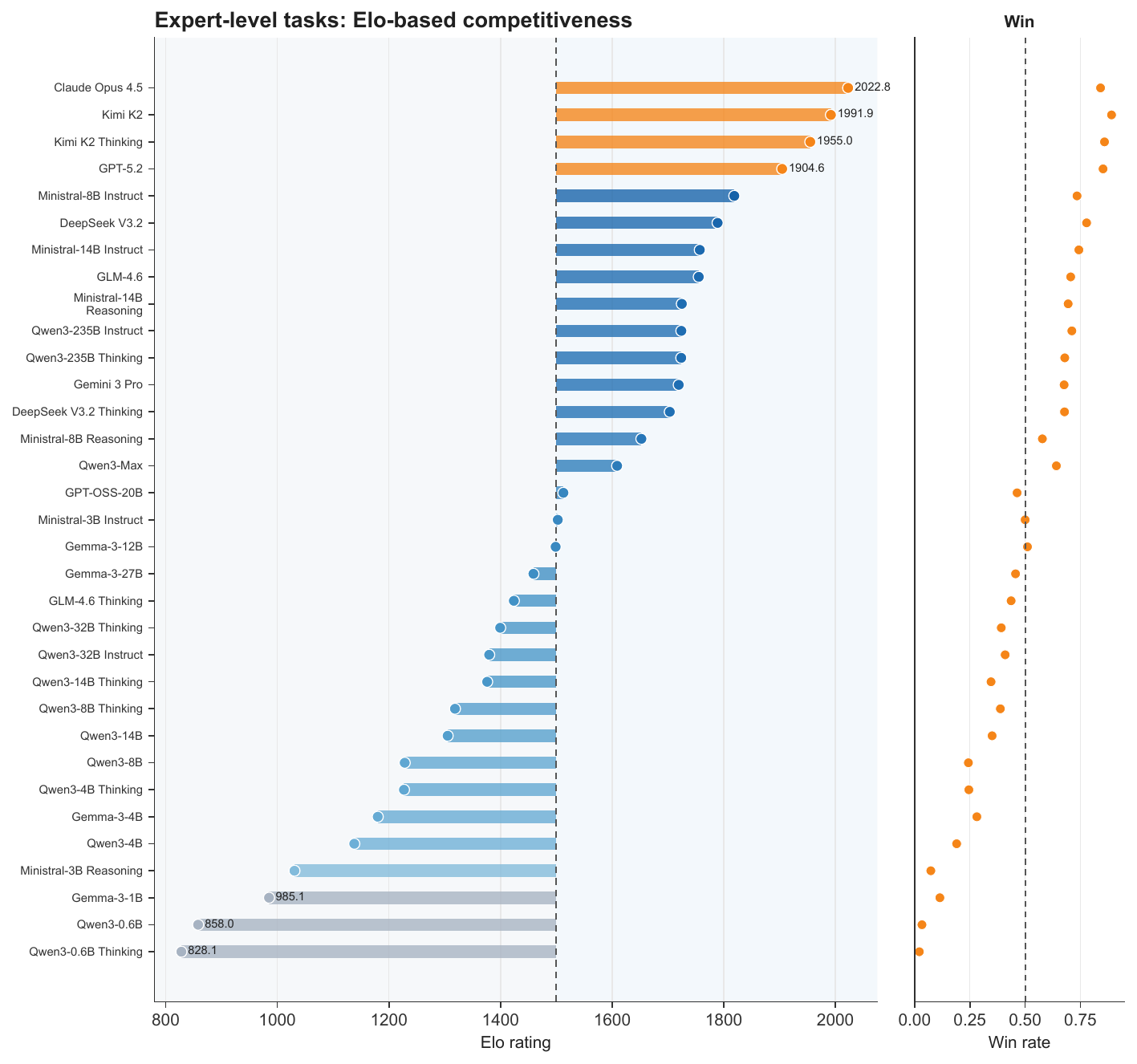}
    \caption{Elo-based evaluation of model competitiveness on expert-level SWM tasks. The main plot shows each model's final Elo score relative to the 1500 initialization point, and the right-side dot panel reports win rates. Top models exhibit both high Elo and stable win rates under the sampled pairwise protocol. Bottom models display limited relative competitiveness, revealing gaps in solution completeness and constraint handling.}
    \label{fig:elo_ranking}
\end{figure}

Figure~\ref{fig:elo_ranking} presents the Elo ranking results for models on expert-level tasks. The main plot orders models from highest to lowest Elo scores, while the right-side panel shows win rates. Overall, differences in relative competitiveness are pronounced. \texttt{claude-opus-4.5} ranks first with an Elo of 2022.8, followed by \texttt{kimi-k2-preview}, \texttt{kimi-k2-thinking}, and \texttt{gpt-5.2}, all exceeding 1900. In contrast, lower-ranked models such as \texttt{Qwen3-0.6B} and \texttt{Qwen3-0.6B-thinking} achieve Elo scores of only 858.1 and 828.1, highlighting their limited relative competitiveness. The gap of over 1100 points between top and bottom models indicates that complex open-ended tasks further amplify hierarchical differences in model capabilities.

Elo scores provide a complementary view of competitive stability. Top models maintain high win rates, while some models with reasonable absolute scores rank lower in Elo, indicating that answer quality and pairwise dominance are related but not identical. Since the stored pairwise protocol yields unequal match counts, match totals are treated as a reproducibility detail rather than a substantive performance indicator.

\subsubsection{Error Analysis}
\label{subsubsec:error_analysis}

\begin{figure}[h]
    \centering
    \includegraphics[width=1\linewidth]{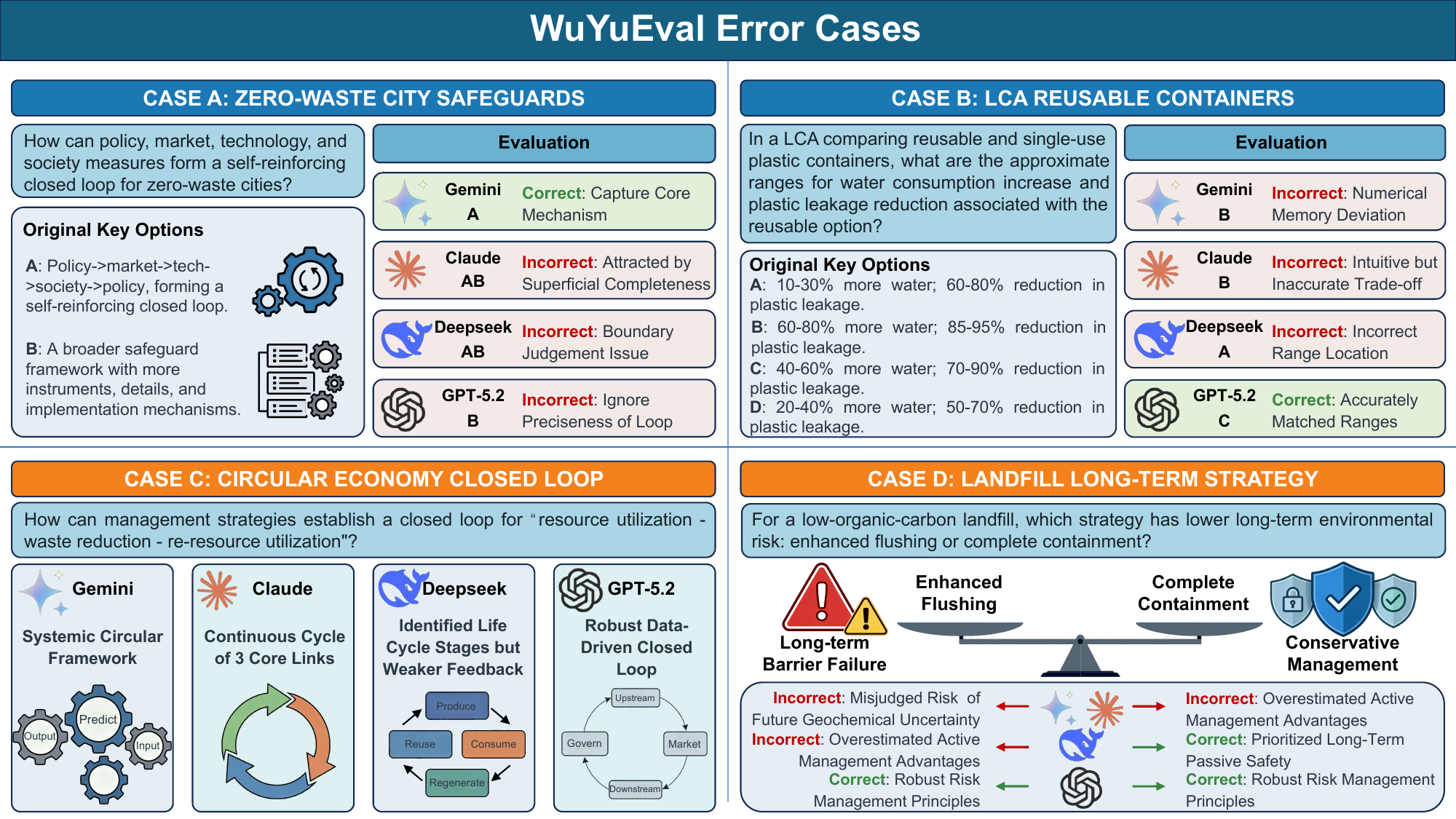}
    \caption{Representative error patterns across WuYuEval tasks. Closed-ended tasks expose surface-completeness bias and instability on quantitative domain facts, while expert-level tasks reveal differences in closed-loop reasoning and long-term risk trade-off judgment.}
    \label{fig:failure_figure}
\end{figure}

Figure~\ref{fig:failure_figure} summarizes four representative error patterns observed in WuYuEval. \textbf{Case A} illustrates \textit{surface-completeness bias} in closed-ended tasks. In a zero-waste city safeguard question, the correct option is the one that explicitly forms a self-reinforcing closed loop across policy, market, technology, society, and back to policy. Gemini selects the correct answer, whereas Claude and DeepSeek select both the correct option and a more detailed distractor, and GPT-5.2 selects the distractor alone. This case shows that models may recognize the relevant elements but fail to enforce the single-best-answer boundary when another option appears richer or more comprehensive.

\textbf{Case B} shows \textit{instability on quantitative domain facts}. In an LCA question about reusable containers, the correct answer depends on matching two numerical intervals: roughly 40--60\% higher water consumption and 70--90\% lower plastic leakage relative to single-use plastic containers. GPT-5.2 selects the correct interval, while Gemini and Claude choose a more extreme trade-off range and DeepSeek chooses a lower range. These errors suggest that models often preserve the qualitative direction of the trade-off, but remain unreliable when the task requires precise recall or localization of domain-specific quantitative values.

\textbf{Case C} concerns \textit{closed-loop organization} in expert-level open-ended tasks. For a circular-economy management question centered on ``resource utilization--waste reduction--re-resource utilization'', strong responses do more than enumerate measures. Gemini and Claude explicitly frame the answer as a systemic circular framework or continuous cycle; GPT-5.2 further connects upstream, midstream, and downstream actions through data governance, accountability mechanisms, and business models. DeepSeek covers the main lifecycle stages, but its response is more stage-wise and less explicit about feedback mechanisms. The difference illustrates a key expert-level distinction: some models list relevant measures, whereas stronger responses explain how these measures reinforce one another as an operational loop.

\textbf{Case D} captures \textit{risk-preference divergence under deep uncertainty}. In a landfill long-term strategy case, the reference answer favors enhanced long-term passive containment or complete containment, because future barrier failure and geochemical uncertainty may make active pollutant mobilization risky. Gemini and Claude instead prefer enhanced flushing, emphasizing inventory reduction and avoidance of a future chemical time bomb. DeepSeek and GPT-5.2 choose complete containment and explicitly connect uncertainty to a conservative risk-management principle. This case highlights that expert SWM evaluation is not only about listing technically plausible options, but also about selecting a defensible risk posture under uncertain long-term environmental behavior.

\subsection{Discussion}
\label{subsec:discussion}

Solid waste management is not only a knowledge domain but also a constraint-coupled decision space. The WuYuEval results show that current LLMs can support parts of this workflow, but their reliability varies strongly with task structure, evidence use, and the degree to which reasoning must remain bounded by engineering conditions.

\paragraph{Implications for SWM applications}

Current models perform best on foundational and normative tasks, such as terminology recognition, regulatory knowledge, and scenario-based judgment with clear answer boundaries. Their weaknesses become more evident in calculation, experimental design, urban planning, and expert-level open-ended tasks, where models must jointly handle numerical constraints, causal relationships, technology selection, environmental impacts, economic feasibility, and policy objectives. In these settings, responses may be fluent and domain-like but still omit key assumptions, misuse evidence, or reach conclusions that are not sufficiently supported by the question conditions. LLMs are therefore better positioned as assistive systems for retrieval, case summarization, preliminary plan drafting, data-gap identification, and comparative textual analysis than as autonomous decision-makers in high-stakes SWM practice.

\paragraph{Reasoning modes and domain logic}

Explicit Thinking modes do not fully solve this problem. The matched-pair comparison suggests that the main bottleneck is not the length of the reasoning trace, but whether the trace remains anchored to reliable evidence, valid units, and engineering constraints. The observed gains are larger for weaker Qwen3 variants, indicating that explicit reasoning can compensate for limited baseline capability by organizing intermediate steps. By contrast, the negative gains for \texttt{GLM-4.6} and \texttt{Kimi-K2} suggest that stronger models may already encode much of the relevant knowledge in standard mode, so additional deliberation can occasionally broaden the decision frame, over-weight plausible narrative steps, or move away from the governing constraint. This pattern should be interpreted as observational rather than causal, because model interfaces and Thinking implementations differ across families. In SWM, correct decisions often depend on a professional logic chain: defining system boundaries, selecting decisive evidence, checking units and mass balances, evaluating feasible technology pathways, and applying environmental, economic, and regulatory constraints.

\paragraph{Implications for foundation-model training}

These results argue for SWM-oriented foundation-model training that couples domain corpora with explicit constraint representations, calculation tools, process templates, and verifiable professional reasoning chains. Rather than encouraging longer generic chain-of-thought outputs, future systems should learn when to retrieve standards, when to calculate, when to preserve uncertainty, when to apply conservative risk principles, and when to defer to expert review.

\paragraph{Limitations}

Several limitations should be considered when interpreting the results. First, the Foundation Module is broad but imbalanced: Basic Knowledge, Scientific Knowledge, and Scenario Analysis account for most questions, whereas Professional Calculation and Experimental Design are smaller subsets. Task-level comparisons should therefore be treated as diagnostic trends rather than precise estimates of domain-wide ability. Second, difficulty labels are calibrated using reference models and manual checks, and thus reflect operational difficulty under this protocol rather than intrinsic task hardness. Third, LLM-as-a-Judge scores depend on the selected judge model, anchor design, and prompt format. The repair script improves robustness by re-judging abnormal files and applying anchor normalization, but it does not replace human expert calibration. Fourth, the comparison between standard and Thinking modes is observational across available model interfaces and does not isolate internal reasoning mechanisms. Finally, future benchmark releases should provide reviewer counts, exclusion rates, inter-reviewer agreement, error-rate statistics, anonymized data and code, and complete license metadata for source materials.

\paragraph{Future directions}

Future work should focus on three directions:
\begin{enumerate}
    \item \textbf{Audited domain corpora}: construct higher-quality SWM QA and scenario datasets with transparent quality-control records.
    \item \textbf{Constraint-aware professional reasoning}: train and evaluate professional reasoning chains that link assumptions, units, mass and energy balances, process feasibility, risk posture, and regulatory constraints, with retrieval, calculators, and program templates where appropriate.
    \item \textbf{Stronger evaluation protocols}: add human-calibrated judging, unit and dimensional checks, causal-chain assessment, blinded model comparison, and structural completeness metrics.
\end{enumerate}

\section{Broader Impact}
\label{sec:broader_impact}

WuYuEval is intended to support more reliable evaluation of LLMs in solid waste management, where incorrect technical or policy recommendations may affect environmental risk control, infrastructure planning, and regulatory compliance. The benchmark can help researchers and practitioners identify model limitations before deployment, especially in calculation, experimental design, urban planning, and expert-level decision-making tasks.

At the same time, WuYuEval should not be interpreted as certifying any model for autonomous professional use. The results show that even strong models can produce plausible but unsupported conclusions, particularly under quantitative uncertainty or multi-objective trade-offs. We therefore recommend using LLMs as assistive tools with traceable data, domain-specific computational checks, and human expert review in high-stakes SWM applications.

\section{Conclusion}
\label{sec:conclusion}

This paper presents WuYuEval, a hierarchical benchmark for evaluating large language models in solid waste management. By combining 4,590 audited foundational multiple-choice questions with 247 expert-level open-ended scenarios, WuYuEval assesses both domain knowledge and higher-order capabilities such as calculation, experimental design, urban planning, and complex decision-making.

The results show that current LLMs have meaningful domain adaptability on basic knowledge, conceptual discrimination, and normatively structured tasks. However, their performance declines substantially in high-difficulty, multi-constraint, and strongly coupled scenarios. Professional calculation, experimental design, urban planning, and expert-level open tasks expose persistent weaknesses in evidence use, constraint integration, structural completeness, and cross-scenario consistency.

A central implication is that visible reasoning must be anchored by professional constraints to be useful. After removing low-quality or low-confidence foundation questions, most matched Thinking modes improve, with larger gains generally appearing in weaker baseline models. However, negative gains in some already strong models suggest that additional deliberation can also lead to over-reasoning or movement away from decisive SWM evidence when it is not constrained by domain logic, units, assumptions, and engineering feasibility. WuYuEval therefore provides an empirical basis for identifying the capability boundaries of current LLMs in SWM and for guiding future SWM-oriented foundation models toward professional reasoning-chain construction, explicit constraint control, and tool-assisted verification.

\section*{Acknowledgments}

We thank the following domain experts for contributing professional knowledge, practical cases, and review feedback during the construction and validation of WuYuEval. The experts are listed in no particular order.

{\small
\begin{longtable}{@{}p{0.27\textwidth}p{0.65\textwidth}@{}}
\toprule
\textbf{Expert} & \textbf{Affiliation} \\
\midrule
\endfirsthead
\toprule
\textbf{Expert} & \textbf{Affiliation} \\
\midrule
\endhead
Cheng Xiaoshi & Hohai University \\
Ren Hang & School of Environment, Tsinghua University \\
Chen Shuyuan & Southwest University of Science and Technology \\
Lin Guannv & School of Environment, Tsinghua University \\
Wang Yadong & School of Environment, Tsinghua University \\
Huang Ziqing & School of Environment, Tsinghua University \\
Lu Jian & Dongguan Qinglv Environmental Technology Co., Ltd. \\
Wang Guojie & Hebei Qingfeng Green Energy Solid Waste Disposal Co., Ltd. \\
Wang Lina & Wu'an Xinfeng Cement Co., Ltd. \\
Liu Bao & Hebei Qingfeng Green Energy Solid Waste Disposal Co., Ltd. \\
Yu Wantong & Wu'an Xinfeng Cement Co., Ltd. \\
Guo Huiping & Wu'an Xinfeng Cement Co., Ltd. \\
Zhao Wenhao & Wu'an Xinfeng Cement Co., Ltd. \\
Xu Zhipeng & School of Environment, Tsinghua University \\
Miao Hua & Wu'an Xinfeng Cement Co., Ltd. \\
Zhang Haixiao & Wu'an Xinfeng Cement Co., Ltd. \\
Li Ying & Wu'an Xinfeng Cement Co., Ltd. \\
Bai Yang & Wu'an Xinfeng Cement Co., Ltd. \\
Zhang Chengfan & Wu'an Xinfeng Cement Co., Ltd. \\
Sun Weiliang & Wu'an Xinfeng Cement Co., Ltd. \\
Gu Xierong & University of Liverpool \\
Yang Chuanxi & Weifang University of Science and Technology \\
Qian Zhifeng & School of Management, Lanzhou University \\
Lu Fei & Hebei Academy of Environmental Sciences \\
Chen Jinyue & Shandong University \\
Wang Shuaima & Jiangxi Academy of Ecological Environment Science and Planning \\
Peng Xiao & China University of Mining and Technology \\
Liu Chenglong & Ningxia Normal University \\
Zhang Weijin & Central South University \\
Lin Kunsen & Fujian Normal University \\
Cheng Zhang & Nanjing Institute of Environmental Sciences, Ministry of Ecology and Environment \\
Huang Liuqing & Nanjing Institute, Ministry of Ecology and Environment \\
Wang Minlong & Institute of Coal Chemistry, Chinese Academy of Sciences, Shanxi \\
Ma Wanqi & Jiangnan University \\
Cai Xingrui & Peking University \\
Shi Huimin & Qifeite Environmental Protection \\
Zheng Zelin & School of Environment, Tsinghua University \\
Chen Jingang & University of Science and Technology Beijing \\
Zhang San & Beijing Zhongqing Huanxun Technology Co., Ltd. \\
Ke Sihua & Beijing Zhongqing Huanxun Technology Co., Ltd. \\
Wei Congxin & Lanzhou University of Technology \\
Wang Bo & Solid Waste and Chemical Management Technology Center, Ministry of Ecology and Environment \\
Zhou Quan & Solid Waste and Chemical Management Technology Center, Ministry of Ecology and Environment \\
Cao Qilin & Jilin University \\
Shi Run & University of Science and Technology Beijing \\
Li Keke & Beijing Zhongqing Huanxun Technology Co., Ltd. \\
Tang Xinru & School of Integrated Circuits, Tsinghua University \\
Zhang Jiale & Tsinghua University \\
Liu Xiaomi & Beijing Zhongqing Huanxun Technology Co., Ltd. \\
Zhou Guokun & Beijing Zhongqing Huanxun Technology Co., Ltd. \\
Liu Wenzhuang & North China University of Science and Technology \\
Wang Chen & Shandong University \\
Wang Yidong & Tsinghua University \\
Yang Zeguo & School of Environment, Huazhong University of Science and Technology \\
Cui Jiaying & Lanzhou University \\
Guan Jinghua & Beijing University of Technology \\
Guo Xin & Tongji University \\
Yao Yuan & Anhui Conch Environmental Protection Group Co., Ltd. \\
Liu Yuanfang & Hubei Junbang Environmental Technology Co., Ltd. \\
Wang Xinzi & Beijing University of Chemical Technology \\
Hu Shuhan & Institute of Geographic Sciences and Natural Resources Research, Chinese Academy of Sciences \\
Zheng Kai & Tsinghua University \\
Huang Junlong & Tsinghua University \\
Xie Dong & School of Environment, Tsinghua University \\
\bottomrule
\end{longtable}
}

\clearpage
\bibliographystyle{plainnat}
\bibliography{refs}

\clearpage
\appendix

\section{WuYuEval Dataset Details}
\label{app:dataset_details}

This appendix provides additional details on the construction and composition of WuYuEval, including dataset statistics, prompt templates used during data construction, and quality-control procedures.

\subsection{Detailed Dataset Statistics}
\label{app:dataset_statistics}

Table~\ref{tab:app_foundation_task_distribution}, Table~\ref{tab:app_foundation_domain_distribution}, and Table~\ref{tab:app_foundation_difficulty_distribution} report the detailed statistics of the Foundation Module. Table~\ref{tab:app_expert_type_distribution} and Table~\ref{tab:app_expert_sublabel_distribution} report the detailed statistics of the Expert Module.

\begin{table}[ht]
\centering
\caption{Distribution of foundation questions by task type.}
\label{tab:app_foundation_task_distribution}
\begin{tabular}{lcc}
\toprule
\textbf{Task Type} & \textbf{Count} & \textbf{Percentage} \\
\midrule
Basic Knowledge & 2,252 & 49.1\% \\
Scientific Knowledge & 610 & 13.3\% \\
Scenario Analysis & 1,380 & 30.1\% \\
Urban Planning & 170 & 3.7\% \\
Experimental Design & 103 & 2.2\% \\
Professional Calculation & 75 & 1.6\% \\
\midrule
Total & 4,590 & 100.0\% \\
\bottomrule
\end{tabular}
\end{table}

\begin{table}[ht]
\centering
\caption{Distribution of foundation questions by topical domain.}
\label{tab:app_foundation_domain_distribution}
\begin{tabular}{lcc}
\toprule
\textbf{Topical Domain} & \textbf{Count} & \textbf{Percentage} \\
\midrule
Waste Treatment and Disposal & 2,474 & 53.9\% \\
Environmental Impact and Risk Assessment & 626 & 13.6\% \\
Circular Economy and Waste-Free Cities & 517 & 11.3\% \\
Management Systems and Standards & 271 & 5.9\% \\
Laws and Regulations & 258 & 5.6\% \\
Solid Waste Fundamentals & 187 & 4.1\% \\
AI Applications in SWM & 119 & 2.6\% \\
Transfer and Collection & 138 & 3.0\% \\
\midrule
Total & 4,590 & 100.0\% \\
\bottomrule
\end{tabular}
\end{table}

\begin{table}[ht]
\centering
\caption{Distribution of foundation questions by difficulty level.}
\label{tab:app_foundation_difficulty_distribution}
\begin{tabular}{lcc}
\toprule
\textbf{Difficulty} & \textbf{Count} & \textbf{Percentage} \\
\midrule
Easy & 3,822 & 83.3\% \\
Medium & 348 & 7.6\% \\
Hard & 420 & 9.2\% \\
\midrule
Total & 4,590 & 100.0\% \\
\bottomrule
\end{tabular}
\end{table}

\begin{table}[ht]
\centering
\caption{Distribution of expert questions by type.}
\label{tab:app_expert_type_distribution}
\begin{tabular}{lcc}
\toprule
\textbf{Expert Type} & \textbf{Count} & \textbf{Percentage} \\
\midrule
General SWM Sub-domains & 88 & 35.6\% \\
Modern Environmental Challenges & 88 & 35.6\% \\
Persistent Environmental Issues & 71 & 28.7\% \\
\midrule
Total & 247 & 100.0\% \\
\bottomrule
\end{tabular}
\end{table}

\begin{table}[ht]
\centering
\caption{Distribution of expert questions by sublabel.}
\label{tab:app_expert_sublabel_distribution}
\begin{tabular}{lcc}
\toprule
\textbf{Sublabel} & \textbf{Count} & \textbf{Percentage} \\
\midrule
Microplastics and Nanoplastics & 34 & 13.8\% \\
Solid Waste Treatment and Disposal Technologies & 33 & 13.4\% \\
General Plastic Waste & 23 & 9.3\% \\
Batteries & 23 & 9.3\% \\
Landfill Remediation & 19 & 7.7\% \\
Wind Turbines & 18 & 7.3\% \\
Comprehensive Queries & 16 & 6.5\% \\
Artificial Intelligence Application of SWM & 15 & 6.1\% \\
Solar Modules & 14 & 5.7\% \\
Circular Economy and Waste-Free Cities & 11 & 4.5\% \\
WEEE & 10 & 4.0\% \\
General Municipal Solid Wastes & 5 & 2.0\% \\
High-Tech Wastes & 5 & 2.0\% \\
Marine Wastes & 4 & 1.6\% \\
Industrial Wastes & 4 & 1.6\% \\
Fundamentals of Solid Waste Management & 3 & 1.2\% \\
Environmental Impact and Risk Assessment & 3 & 1.2\% \\
Solid Waste Transfer and Collection & 3 & 1.2\% \\
Solid Waste Management Systems and Standards & 2 & 0.8\% \\
Solid Waste Laws, Regulations and Policies & 2 & 0.8\% \\
\midrule
Total & 247 & 100.0\% \\
\bottomrule
\end{tabular}
\end{table}

\subsection{Prompt Templates for Data Construction}
\label{app:construction_prompts}

This section provides representative prompt templates used during the construction of WuYuEval. These prompts are used to convert multi-source SWM materials into standardized questions, expert-level open-ended tasks, and reference solution elements.

\subsubsection{Prompt for Basic Knowledge Questions}
\label{app:prompt_knowledge}

\begin{tcolorbox}[title=Prompt for Basic Knowledge]
You are an expert in solid waste management. Given the following professional material, extract atomic knowledge points and generate multiple-choice questions that test conceptual understanding.

Requirements:
1. Each question should focus on one clear knowledge point.
2. The question should have exactly one correct option supported by the provided evidence.
3. Distractors should be plausible but scientifically incorrect.
4. Avoid ambiguous wording or options that can be interpreted as partially correct.
5. Provide the correct answer and a concise explanation.

Input material:
\{material\}

Output format:
Question:
Options:
A.
B.
C.
D.
Answer:
Explanation:
Domain label:
Difficulty:
\end{tcolorbox}

\subsubsection{Prompt for Scenario Analysis Questions}
\label{app:prompt_scenario}

\begin{tcolorbox}[title=Prompt for Scenario Analysis]
You are an environmental engineering expert specializing in solid waste management. Based on the given case material, construct a scenario-based multiple-choice question.

The question should require the model to reason under explicit constraints, such as waste type, treatment objective, infrastructure condition, environmental risk, economic feasibility, or policy requirement.

Requirements:
1. The scenario should be realistic and technically grounded.
2. The options should require comparison or causal reasoning rather than simple recall.
3. The correct answer must be supported by the input material.
4. Include a domain label, source type, and difficulty level.

Input case:
\{case\}

Output format:
Question:
Options:
A.
B.
C.
D.
Answer:
Explanation:
Domain label:
Difficulty:
\end{tcolorbox}

\subsubsection{Prompt for Professional Calculation Questions}
\label{app:prompt_calculation}

\begin{tcolorbox}[title=Prompt for Professional Calculation]
You are an expert in environmental engineering and solid waste management. Generate a professional calculation question based on the given calculation kernel.

Requirements:
1. Preserve the original engineering logic.
2. Randomize numerical variables within realistic ranges.
3. Ensure that the question has a unique and verifiable answer.
4. Provide step-by-step calculation reasoning and the final answer.
5. Avoid values that lead to numerical ambiguity or unrealistic engineering conditions.

Calculation kernel:
\texttt{\{calculation\_kernel\}}

Output format:
Question:
Options:
A.
B.
C.
D.
Answer:
Calculation process:
Domain label:
Difficulty:
\end{tcolorbox}

\subsubsection{Prompt for Scientific Knowledge Questions}
\label{app:prompt_scientific}

\begin{tcolorbox}[title=Prompt for Scientific Knowledge]
You are a researcher in solid waste management. Given the following academic abstract or technical passage, identify the core scientific conclusions and transform them into rigorous multiple-choice questions.

Requirements:
1. Focus on mechanisms, experimental findings, technical principles, or causal relationships.
2. Avoid questions that only test memorization of isolated facts.
3. Ensure that each correct option is directly supported by the input passage.
4. Provide concise explanations for both correct and incorrect options.

Input passage:
\{paper\_excerpt\}

Output format:
Question:
Options:
A.
B.
C.
D.
Answer:
Explanation:
Research topic:
Domain label:
Difficulty:
\end{tcolorbox}

\subsubsection{Prompt for Expert-Level Questions}
\label{app:prompt_expert}

\begin{tcolorbox}[title=Prompt for Expert-Level Open-ended Questions]
You are a senior expert in solid waste management. Based on the following real-world engineering problem, construct an expert-level open-ended evaluation question.

The question should require system-level reasoning and should not have a single unique answer. It should involve multiple constraints, such as technical feasibility, environmental risk, economic cost, infrastructure condition, policy compliance, and long-term sustainability.

Requirements:
1. Clearly describe the scenario and decision context.
2. Include multiple interacting constraints.
3. Require the model to propose a feasible and coherent solution.
4. Provide reference solution elements rather than a single standard answer.
5. Include evaluation dimensions for judge-based scoring.

Input expert material:
\{expert\_material\}

Output format:
Question:
Reference solution elements:
Key constraints:
Expected reasoning dimensions:
Evaluation rubric:
Expert type:
Sublabel:
\end{tcolorbox}

\subsubsection{Prompt for LLM-as-a-Judge Overall Scoring}
\label{app:judge_prompts}

\begin{tcolorbox}[title=Prompt for Overall Judge Scoring]
You are scoring model answers absolutely versus the golden answer. The input contains the question, the golden answer, all model answers, and two anchors: \texttt{\_\_GOLD\_\_} and \texttt{\_\_NULL\_\_}. Return only one short JSON object without markdown:
\[
\{\texttt{"model\_scores"}:[\{\texttt{"model\_name"}:\texttt{"name"},\texttt{"raw\_score"}:0\}]\}.
\]
Rules: \texttt{raw\_score} must be in [0,100], and all shown models, including both anchors, must be included.
\end{tcolorbox}

\subsubsection{Prompt for Answer-Graph Structural Scoring}
\label{app:graph_judge_prompt}

\begin{tcolorbox}[title=Prompt for Graph Judge Scoring]
You are scoring knowledge graphs absolutely versus the golden answer. The input contains the question, the golden answer, all model graphs, and two anchors: \texttt{\_\_GOLD\_\_} and \texttt{\_\_NULL\_\_}. Return only one short JSON object without markdown:
\[
\{\texttt{"model\_graph\_scores"}:[\{\texttt{"model\_name"}:\texttt{"name"},\texttt{"raw\_graph\_score"}:0\}]\}.
\]
Rules: \texttt{raw\_graph\_score} must be in [0,100], and all shown models, including both anchors, must be included.
\end{tcolorbox}

\subsubsection{Prompt for Elo Pairwise Comparison}
\label{app:elo_judge_prompt}

\begin{tcolorbox}[title=Prompt for Pairwise Elo Judgment]
You are judging a head-to-head comparison between two model outputs for the same question. Judge which output is better relative to the golden answer, considering both answer quality and graph quality. Return JSON only with \texttt{winner} as A, B, or TIE, \texttt{confidence} as high, medium, or low, and a one-sentence reason. Penalize major factual errors and fabrication heavily.
\end{tcolorbox}

\section{Full Experimental Results}
\label{app:full_results}

This appendix reports the complete evaluation results of all evaluated models on WuYuEval. Section~\ref{app:foundation_results} presents results on the Foundation Module, and Section~\ref{app:expert_results} presents results on the Expert Module.

\subsection{Foundation Module Results}
\label{app:foundation_results}

\subsubsection{Overall Model Performance}
\label{app:foundation_overall}

\begin{table}[h]
\centering
\caption{Overall performance of the 33 evaluated models on the Foundation Module. \texttt{Qwen3-14B-sft} is excluded from all reported statistics.}
\label{tab:overall_performance}
\begin{tabularx}{\textwidth}{>{\raggedright\arraybackslash}p{0.6\textwidth}
                            >{\centering\arraybackslash}X
                            >{\centering\arraybackslash}X}
\toprule
\textbf{Model} & \textbf{ACC} & \textbf{Macro-F1} \\
\midrule
claude-opus-4.5 & 94.64\% & 95.73\% \\
gemini-3-pro-preview & 92.77\% & 95.53\% \\
glm-4.6 & 92.31\% & 93.54\% \\
DeepSeek-V3.2-Thinking & 92.16\% & 93.51\% \\
kimi-k2-preview & 92.00\% & 92.85\% \\
qwen3-max & 91.83\% & 93.09\% \\
DeepSeek-V3.2 & 91.37\% & 92.66\% \\
kimi-k2-thinking & 91.26\% & 92.23\% \\
qwen3-235b-a22b-thinking & 90.83\% & 92.54\% \\
gpt-5.2 & 90.04\% & 91.36\% \\
qwen3-235b-a22b-instruct & 89.63\% & 90.93\% \\
qwen3-32b-thinking & 87.86\% & 89.58\% \\
qwen3-32b-instruct & 87.49\% & 89.18\% \\
glm-4.6-thinking & 84.97\% & 86.53\% \\
gemma-3-27b-it & 82.29\% & 84.73\% \\
Qwen3-14B-thinking & 81.37\% & 87.70\% \\
ministral-3-14b-instruct & 81.00\% & 84.71\% \\
ministral-3-8b-instruct & 79.74\% & 82.91\% \\
ministral-3-14b-reasoning & 79.24\% & 82.23\% \\
gpt-oss-20b & 77.60\% & 78.01\% \\
Qwen3-14B & 76.47\% & 83.09\% \\
ministral3-3b-ins-local & 75.97\% & 78.80\% \\
Qwen3-4B-thinking & 74.75\% & 79.07\% \\
Qwen3-4B & 73.86\% & 77.52\% \\
gemma-3-12b-it & 73.36\% & 79.24\% \\
ministral-3-8b-reasoning & 71.87\% & 79.18\% \\
ministral3-3b-reasoning-local & 68.78\% & 73.30\% \\
gemma-3-4b-it & 65.56\% & 71.12\% \\
Qwen3-8B-thinking & 64.40\% & 77.22\% \\
Qwen3-8B & 62.88\% & 75.70\% \\
gemma-3-1b-it & 46.99\% & 57.82\% \\
Qwen3-0.6B-thinking & 44.99\% & 61.98\% \\
Qwen3-0.6B & 37.86\% & 53.94\% \\
\bottomrule
\end{tabularx}
\end{table}

\subsubsection{Model performance on the Foundation Module by difficulty level.}
\label{app:foundation_difficulty_results}

\begin{table}[ht]
\centering
\caption{Model performance by difficulty level. $N$ denotes model--question evaluation records across the 33 evaluated models.}
\label{tab:difficulty}
\begin{tabularx}{\textwidth}{>{\raggedright\arraybackslash}X
                             >{\centering\arraybackslash}X
                             >{\centering\arraybackslash}X
                             >{\centering\arraybackslash}X}
\toprule
\textbf{Difficulty} & \textbf{N} & \textbf{ACC} & \textbf{Macro-F1} \\
\midrule
Easy & 126126 & 84.14\% & 88.30\% \\
Medium & 11484 & 59.10\% & 64.06\% \\
Hard & 13860 & 42.50\% & 49.98\% \\
\bottomrule
\end{tabularx}
\end{table}

\subsubsection{Performance by Task Type}
\label{app:foundation_task_results}

\begin{table}[ht]
\centering
\caption{Model performance on the Foundation Module by task type.}
\label{tab:task_types}
\begin{tabularx}{\textwidth}{>{\raggedright\arraybackslash}p{0.5\textwidth}
                             >{\centering\arraybackslash}X
                             >{\centering\arraybackslash}X}
\toprule
\textbf{Source} & \textbf{ACC} & \textbf{Macro-F1} \\
\midrule
Basic Knowledge & 85.52\% & 87.69\% \\
Scientific Knowledge & 70.50\% & 72.72\% \\
Scenario Analysis & 78.31\% & 88.10\% \\
Urban Planning & 47.31\% & 50.10\% \\
Experimental Design & 41.84\% & 44.65\% \\
Professional Calculation & 52.97\% & 56.34\% \\
\bottomrule
\end{tabularx}
\end{table}

\paragraph{Domain analysis.}
Table~\ref{tab:domains} presents performance across SWM domains. Basic Concepts and Classification achieves the highest ACC, while Circular Economy and Waste-Free Cities is among the most challenging categories. The latter requires models to integrate technical, economic, policy, and system-level considerations, making it substantially more difficult than isolated knowledge recognition.

\subsubsection{Performance by Topical Domain}
\label{app:foundation_domain_results}

\begin{table}[ht]
\centering
\caption{Model performance on the Foundation Module by topical domain.}
\label{tab:domains}
\begin{tabularx}{\textwidth}{>{\raggedright\arraybackslash}p{0.55\textwidth}
                             >{\centering\arraybackslash}X
                             >{\centering\arraybackslash}X}
\toprule
\textbf{Label} & \textbf{ACC} & \textbf{Macro-F1} \\
\midrule
Solid Waste Treatment and Disposal Technologies & 79.97\% & 83.87\% \\
Environmental Impact and Risk Assessment of Solid Waste & 77.32\% & 82.88\% \\
Circular Economy and Waste-Free Cities & 67.04\% & 73.97\% \\
Solid Waste Management System and Standards & 79.19\% & 83.57\% \\
Solid Waste Laws, Regulations, and Policies & 83.25\% & 87.82\% \\
Basic Concepts and Classification of Solid Waste & 85.01\% & 89.06\% \\
Artificial Intelligence Applications in SWM & 80.34\% & 83.89\% \\
Solid Waste Transfer and Collection & 77.32\% & 81.29\% \\
\bottomrule
\end{tabularx}
\end{table}

\subsubsection{Performance of Thinking and Non-thinking Modes}
\label{app:thinking_mode_results}

\begin{table}[ht]
\centering
\small
\renewcommand{\arraystretch}{1.15}
\setlength{\tabcolsep}{6pt}
\caption{Comparison between standard and thinking modes on the Foundation Module. Models are ordered by $\Delta$ACC. All values are reported in percentage points.}
\label{tab:app_thinking_mode_results}
\begin{tabular}{lrrrrrr}
\toprule
\textbf{Model Family} 
& \textbf{ACC$_{\mathrm{standard}}$} 
& \textbf{ACC$_{\mathrm{thinking}}$} 
& \textbf{$\Delta$ACC} 
& \textbf{F1$_{\mathrm{standard}}$} 
& \textbf{F1$_{\mathrm{thinking}}$} 
& \textbf{$\Delta$F1} \\
\midrule
GLM-4.6            & 92.31 & 84.97 & -7.34 & 93.54 & 86.53 & -7.02 \\
Kimi-K2            & 92.00 & 91.26 & -0.74 & 92.85 & 92.23 & -0.62 \\
Qwen3-32B          & 87.49 & 87.86 & +0.37 & 89.18 & 89.58 & +0.41 \\
DeepSeek-V3.2      & 91.37 & 92.16 & +0.78 & 92.66 & 93.51 & +0.85 \\
Qwen3-4B           & 73.86 & 74.75 & +0.89 & 77.52 & 79.07 & +1.56 \\
Qwen3-235B-A22B    & 89.63 & 90.83 & +1.20 & 90.93 & 92.54 & +1.60 \\
Qwen3-8B           & 62.88 & 64.40 & +1.53 & 75.70 & 77.22 & +1.52 \\
Qwen3-14B          & 76.47 & 81.37 & +4.90 & 83.09 & 87.70 & +4.61 \\
Qwen3-0.6B         & 37.86 & 44.99 & +7.12 & 53.94 & 61.98 & +8.04 \\
\bottomrule
\end{tabular}
\end{table}

\subsection{Expert Module Results}
\label{app:expert_results}

\subsubsection{LLM-as-a-Judge Scores}
\label{app:judge_score_results}

\begin{table}[ht]
\centering
\caption{LLM-as-a-Judge scores for expert-level tasks.}
\label{tab:judge_score}
\begin{tabularx}{\textwidth}{>{\raggedright\arraybackslash}p{0.45\textwidth}
                             >{\centering\arraybackslash}X
                             >{\centering\arraybackslash}X
                             >{\centering\arraybackslash}X
                             >{\centering\arraybackslash}X}
\toprule
\textbf{Model} & \textbf{Mean} & \textbf{Std} & \textbf{Max} & \textbf{Min} \\
\midrule
qwen3-235b-a22b-thinking & 0.8729 & 0.0939 & 0.9811 & 0.2747 \\
qwen3-max & 0.8708 & 0.0863 & 0.9747 & 0.2950 \\
gpt-5.2 & 0.8646 & 0.0879 & 0.9747 & 0.4000 \\
kimi-k2-thinking & 0.8560 & 0.0923 & 0.9716 & 0.2800 \\
kimi-k2-preview & 0.8518 & 0.0944 & 0.9779 & 0.3300 \\
qwen3-235b-a22b-instruct & 0.8514 & 0.1016 & 0.9716 & 0.2250 \\
claude-opus-4.5 & 0.8422 & 0.1300 & 0.9555 & 0.0000 \\
gemini-3-pro-preview & 0.8242 & 0.1126 & 0.9526 & 0.1800 \\
glm-4.6 & 0.8093 & 0.0974 & 0.9589 & 0.2300 \\
qwen3-32b-thinking & 0.8025 & 0.1230 & 0.9589 & 0.0000 \\
Qwen3-14B-thinking & 0.8012 & 0.0923 & 0.9684 & 0.4118 \\
DeepSeek-V3.2-Thinking & 0.7983 & 0.1240 & 0.9589 & 0.0158 \\
qwen3-32b-instruct & 0.7869 & 0.1153 & 0.9653 & 0.2400 \\
Ministral-3-14B-Instruct & 0.7850 & 0.1128 & 0.9400 & 0.3205 \\
DeepSeek-V3.2 & 0.7833 & 0.1022 & 0.9684 & 0.3295 \\
Qwen3-14B & 0.7825 & 0.0947 & 0.9621 & 0.3700 \\
Ministral-3-14B-Reasoning & 0.7755 & 0.1053 & 0.9558 & 0.3261 \\
Qwen3-8B-thinking & 0.7648 & 0.0976 & 0.9495 & 0.3708 \\
glm-4.6-thinking & 0.7545 & 0.1130 & 0.9653 & 0.2400 \\
gemma-3-27b-it & 0.7531 & 0.1204 & 0.9621 & 0.2471 \\
Ministral-3-8B-Instruct & 0.7526 & 0.1085 & 0.9463 & 0.2595 \\
Qwen3-8B & 0.7492 & 0.1005 & 0.9463 & 0.3200 \\
Ministral-3-8B-Reasoning & 0.7303 & 0.1111 & 0.9495 & 0.2437 \\
gpt-oss-20b & 0.7271 & 0.1403 & 0.9495 & 0.0000 \\
gemma-3-12b-it & 0.7244 & 0.1215 & 0.9463 & 0.2279 \\
Qwen3-4B-thinking & 0.7062 & 0.1099 & 0.9526 & 0.2400 \\
Qwen3-4B & 0.6951 & 0.1115 & 0.9432 & 0.2050 \\
Ministral3-3b-ins & 0.6666 & 0.1218 & 0.9337 & 0.2245 \\
gemma-3-4b-it & 0.6496 & 0.1339 & 0.9432 & 0.1650 \\
Ministral3-3b-reasoning & 0.5879 & 0.1297 & 0.9274 & 0.1900 \\
gemma-3-1b-it & 0.5068 & 0.1385 & 0.8200 & 0.1250 \\
Qwen3-0.6B-thinking & 0.4992 & 0.1322 & 0.9305 & 0.1100 \\
Qwen3-0.6B & 0.4703 & 0.1350 & 0.9211 & 0.0850 \\
\bottomrule
\end{tabularx}
\end{table}

\subsubsection{Elo-based Pairwise Ranking}
\label{app:elo_results}

\begin{table}[ht]
\centering
\caption{Elo rankings of models on expert-level tasks.}
\label{tab:elo_ranking}
\begin{tabularx}{\textwidth}{>{\raggedright\arraybackslash}X
                             >{\centering\arraybackslash}c
                             >{\centering\arraybackslash}c
                             >{\centering\arraybackslash}c}
\toprule
\textbf{Model} & \textbf{Elo} & \textbf{Matches} & \textbf{Win Rate} \\
\midrule
claude-opus-4.5 & 2022.80 & 176 & 0.8409 \\
kimi-k2-preview & 1991.94 & 192 & 0.8906 \\
kimi-k2-thinking & 1954.97 & 170 & 0.8588 \\
gpt-5.2 & 1904.58 & 176 & 0.8523 \\
Ministral-3-8B-Instruct & 1818.82 & 177 & 0.7345 \\
DeepSeek-V3.2 & 1788.91 & 180 & 0.7778 \\
Ministral-3-14B-Instruct & 1756.99 & 175 & 0.7429 \\
glm-4.6 & 1755.15 & 170 & 0.7059 \\
Ministral-3-14B-Reasoning & 1724.81 & 180 & 0.6944 \\
qwen3-235b-a22b-instruct & 1723.93 & 180 & 0.7111 \\
qwen3-235b-a22b-thinking & 1723.74 & 187 & 0.6791 \\
gemini-3-pro-preview & 1719.25 & 176 & 0.6761 \\
DeepSeek-V3.2-Thinking & 1703.23 & 177 & 0.6780 \\
Ministral-3-8B-Reasoning & 1652.33 & 180 & 0.5778 \\
qwen3-max & 1608.84 & 170 & 0.6412 \\
gpt-oss-20b & 1512.59 & 192 & 0.4635 \\
Ministral3-3b-ins & 1502.52 & 184 & 0.5000 \\
gemma-3-12b-it & 1498.67 & 190 & 0.5105 \\
gemma-3-27b-it & 1458.88 & 173 & 0.4566 \\
glm-4.6-thinking & 1423.80 & 181 & 0.4365 \\
qwen3-32b-thinking & 1399.30 & 176 & 0.3920 \\
qwen3-32b-instruct & 1379.60 & 188 & 0.4096 \\
Qwen3-14B-thinking & 1375.99 & 188 & 0.3457 \\
Qwen3-8B-thinking & 1318.31 & 183 & 0.3880 \\
Qwen3-14B & 1305.19 & 174 & 0.3506 \\
Qwen3-8B & 1228.24 & 189 & 0.2434 \\
Qwen3-4B-thinking & 1226.91 & 167 & 0.2455 \\
gemma-3-4b-it & 1179.93 & 174 & 0.2816 \\
Qwen3-4B & 1137.84 & 184 & 0.1902 \\
Ministral3-3b-reasoning & 1030.69 & 177 & 0.0734 \\
gemma-3-1b-it & 985.12 & 175 & 0.1143 \\
Qwen3-0.6B & 858.05 & 180 & 0.0333 \\
Qwen3-0.6B-thinking & 828.07 & 187 & 0.0214 \\
\bottomrule
\end{tabularx}
\end{table}

\section{Additional Qualitative Examples}
\label{app:case_visualization}

This appendix provides additional qualitative examples corresponding to the error analysis in Section~\ref{subsubsec:error_analysis}. The examples illustrate how WuYuEval distinguishes answer-boundary control, quantitative factual precision, closed-loop reasoning, and long-term risk judgment.

\subsection{Foundation Module Examples}
\label{app:foundation_examples}

\paragraph{Case A: zero-waste city safeguards.}
This multiple-choice question asks which safeguard best forms a self-reinforcing closed loop among policy, market, technology, society, and policy feedback. The correct answer is A. Gemini selects A, indicating that it identifies the core loop structure. Claude and DeepSeek select AB: they appear to recognize the correct loop in A, but are also attracted by the richer and more detailed content of B. GPT-5.2 selects B, favoring a comprehensive-looking answer that does not correspond as precisely to the self-reinforcing closed-loop mechanism. This example shows how models can be distracted by surface completeness in single-answer questions.

\paragraph{Case B: LCA of reusable containers.}
This multiple-choice question tests two quantitative intervals: reusable containers cause approximately 40--60\% higher water consumption while reducing plastic leakage by approximately 70--90\% compared with single-use plastic containers. The correct answer is C. GPT-5.2 selects C, while Gemini and Claude select B, a more extreme trade-off range with higher water consumption and higher leakage reduction. DeepSeek selects A, which gives lower ranges for both quantities. The case suggests that models often capture the qualitative direction of an LCA trade-off but remain unstable on specific numerical ranges.

\subsection{Expert Module Examples}
\label{app:expert_examples}

\paragraph{Case C: circular-economy closed loop.}
This expert-level task asks models to design management strategies around the cycle of resource utilization, waste reduction, and re-resource utilization. Gemini frames the answer as a shift from a linear approach to a systemic circular framework and discusses input-stage measures, eco-design, and internal recycling loops. Claude directly describes the three core links as a continuous cycle and expands on material flow analysis and supplier integration. DeepSeek covers resource utilization, production and consumption, waste reduction, and resource regeneration, but its organization is more stage-wise and less explicit about feedback. GPT-5.2 provides a more integrated answer by linking upstream, midstream, and downstream actions through data governance, accountability mechanisms, and business models. The example illustrates that expert performance depends not only on coverage of measures, but also on explaining how measures connect into a functioning loop.

\paragraph{Case D: landfill long-term strategy.}
This expert-level task concerns long-term landfill strategy under uncertainty about future barrier failure and metal migration. The reference answer favors enhanced long-term passive containment or complete containment, because a conservative passive-safety strategy may reduce overall environmental risk under deep geochemical uncertainty. Gemini and Claude instead prefer enhanced flushing, emphasizing pollutant inventory reduction and avoidance of a future chemical time bomb. Their answers capture the logic of active stabilization but give less weight to long-term operation, pollutant mobilization, and uncertainty risks. DeepSeek and GPT-5.2 choose complete containment; GPT-5.2 explicitly argues that avoiding active pollutant mobilization is more robust under uncertainty. This example shows how expert-level tasks reveal differences in risk posture, not merely differences in technical vocabulary.

\clearpage

\end{document}